\documentclass[11pt]{article}
 \usepackage{benstyle}
 
\usepackage{enumerate}

\begin{document}

\title{A Unified Framework for Learning with Nonlinear Model Classes from Arbitrary Linear Samples}

\author{Ben Adcock, ben\_adcock@sfu.ca \\
       Department of Mathematics \\
       Simon Fraser University\\
       Burnaby, BC, Canada
       \and
       Juan M.\ Cardenas,  juan.cardenas@pucv.cl \\
       Instituto de Matem{\'a}ticas\\
		Pontificia Universidad Cat{\'o}lica de Valpara{\'i}so\\
		Valpara{\'i}so, Chile
       \and
       Nick Dexter, nick.dexter@fsu.edu \\
      Department of Scientific Computing \\
       Florida State University \\
      Tallahassee, FL, USA}

\maketitle

\begin{abstract}
We study the fundamental problem of learning an unknown object from data using a prescribed model class. We introduce a unified framework that accommodates objects in arbitrary Hilbert spaces, general (possibly vector-valued) random linear measurements and general types of nonlinear models.
We establish novel learning guarantees for this framework that explicitly relate the required amount of data to structural properties of the model class, yielding near-optimal generalization bounds. A central concept we introduce is the \textit{variation} of a model class relative to a distribution of sampling operators, which quantifies how the model interacts with the measurement process. Combined with entropy integrals that capture the model's complexity, this forms the foundation of our guarantees. Our framework is sufficiently general to recover and unify various well-known problems, such as matrix sketching, compressed sensing with isotropic measurements and compressed sensing with generative models. In each case, existing results arise as direct corollaries of our theory. For compressed sensing with generative models, we also derive the first guarantees for arbitrary Lipschitz generative maps combined with general linear measurements. Overall, our work provides a unified perspective on learning from general data and introduces novel theoretical guarantees that consolidate, sharpen and extend existing results.
\end{abstract}

\section{Introduction}\label{s:introduction}

Learning an unknown object (e.g., a vector, matrix or function) from a finite set of training data is a fundamental problem in applied mathematics and computer science. Typically, in modern settings, one seeks to learn an approximate representation in a nonlinear model class (also known as {an} approximation space or hypothesis set). It is also common to generate the training data randomly according to some distribution. Of critical importance in this endeavour is the question of learning guarantees. In other words: \textit{how much training data suffices to ensure good generalization, and how is this influenced by, firstly, the choice of model class, and secondly, the random process generating the training data?}

This question has often been addressed for specific types of training data. For instance, in the case of function regression, the training data is pointwise evaluations of some target function, or in the case of computational imaging, the training data may consist of samples of the Fourier or Radon transform of some target image. It is also commonly studied for specific model classes, e.g., (linear or nonlinear) polynomial spaces  or (nonlinear) spaces of sparse vectors \cite{foucart2013mathematical,vidyasagar2019introduction}, spaces of low-rank matrices or tensors \cite{davenport2016overview,vidyasagar2019introduction}, spaces defined by generative models \cite{bora2017compressed}, single-  or multi-layer neural networks \cite{gajjar2023active,adcock2021deep,adcock2025nearoptimal}, Fourier sparse functions \cite{erdelyi2020fourier}, (sparse) random feature models \cite{avron2017random,hashemi2023generalization} and many more. 

In this paper, we introduce a unified framework for learning in arbitrary model classes from general linear measurements. Its main features are as follows:
\begin{enumerate}[(i)]
\item the target object is an element of a separable Hilbert space;
\item each measurement is taken randomly and independently according to a random linear operator (which may differ from measurement to measurement);
\item the measurements may be scalar- or vector-valued, or, in general, may take values in an infinite-dimensional Hilbert space;
\item the measurements can be completely \textit{multimodal}, i.e., generated by different distributions of random linear operators, as long as a certain \textit{nondegeneracy} condition holds;
\item the model class can be linear or nonlinear;
\item the resulting learning guarantees are given in terms of the \textit{variation} of the model class with respect to the sampling distributions, plus certain \textit{entropy integrals}.
\end{enumerate}
Variation is a key concept that we introduce in this work, which describes how the model class interacts with the measurements. In tandem with suitable entropy integrals, which describe the complexity of the model class, it lies at the heart of our learning guarantees. After introducing these guarantees, we then present a series of examples to highlight the generality of this framework. In various cases, our resulting guarantees either include or improve known results in the literature.

\subsection{The framework}\label{s:setup}

The setup we consider in this paper is the following.

\begin{itemize}
\item $\bbX$ is a separable Hilbert space with inner product $\ip{\cdot}{\cdot}_{\bbX}$ and norm $\nms{\cdot}_{\bbX}$ and $\bbX_0 \subseteq \bbX$ is a seminormed vector space, termed the \textit{object space}, with seminorm $\nms{\cdot}_{\bbX_0}$. 
\item  $x \in \bbX_0$ is the unknown target object.
\item For each $i = 1,\ldots,m$, $\bbY_i$ is a Hilbert space with inner product $\ip{\cdot}{\cdot}_{\bbY_i}$ and norm $\nms{\cdot}_{\bbY_i}$ termed the $i$th  \textit{measurement space}.
\item For each $i = 1,\ldots,m$, $\cA_i$ is a distribution of bounded linear operators $(\bbX_0,\nms{\cdot}_{\bbX_0} ) \rightarrow (\bbY_i , \nms{\cdot}_{\bbY_i})$. We term $A_i \sim \cA_i$ the $i$th \textit{sampling operator}. We also write $\cB(\bbX_0,\bbY_i)$ for the space of bounded linear operators, so that $A_i \in \cB(\bbX_0,\bbY_i)$.
\item We assume that the family $\{ \cA_i \}^{m}_{i=1}$ of distributions is \textit{nondegenerate}. Namely, there exist $0 < \alpha \leq \beta < \infty$ such that
\be{
\label{nondegeneracy}
\alpha \nm{x}^2_{\bbX} \leq \frac1m \sum^m_{i=1} \bbE_{A_i \sim \cA_i} \nm{A_i(x)}^2_{\bbY_i} \leq \beta \nm{x}^2_{\bbX},\quad \forall x \in \bbX_0.
}
If \ef{nondegeneracy} holds with $\alpha = \beta = 1$, then we say the family $\{ \cA_i \}^{m}_{i=1}$ is \textit{isotropic}.

\item $\bbU \subseteq \bbX_0$ is a set, termed the \textit{approximation space} or \textit{model class}. Our aim is to learn $x \in \bbX_0$ from its measurements with an element $\hat{x} \in \bbU$.
\end{itemize}
Now let $A_i \in \cB(\bbX_0,\bbY_i)$, $i = 1,\ldots,m$, be independent realizations from the distributions $\cA_1,\ldots,\cA_m$. Then we consider noisy measurements
\be{
\label{noisy-meas}
b_i = A_i(x) + e_i \in \bbY_i,\quad i = 1,\ldots,m.
}
In other words, we consider training data of the form
\bes{
\{ (A_i , b_i ) \}^{m}_{i=1},\qquad \text{where }(A_i,b_i) \in \cB(\bbX_0,\bbY_i) \times \bbY_i.
}
Our aim is to recover $x$ from this data using an element of $\bbU$. We do this via the empirical least squares. That is, we let
\be{
\label{least-squares-problem}
\hat{x} \in \argmin{u \in \bbU} \frac1m \sum^{m}_{i=1} \nm{b_i - A_i(u)}^2_{\bbY_i}.
}
Later, we also allow for $\hat{x}$ to be an approximate minimizer, to model the scenario where the minimization problem is {solved} inexactly. Note that in this work we consider \textit{agnostic learning}, where $x \notin \bbU$ and the noise $e_i$ can be adversarial (but small in norm).

As we explain in \S \ref{s:examples}, this framework contains many problems of interest. In particular, scalar-valued and vector-valued function regression from i.i.d.\ samples, matrix sketching for large least-squares problems, compressed sensing with isotropic vectors and compressed sensing with subsampled unitary matrices are all instances of this framework. This includes classical compressed sensing with sparse vectors, but also many generalizations to structured sparse models and, in particular, so-called \textit{compressed sensing with generative models}  \cite{bora2017compressed,berk2022coherence,jalal2021robust}.

As we discuss in \S \ref{s:examples}, it is often the case that $\cA_1 = \cdots = \cA_m = \cA$. However, having different distributions allows us to consider \textit{multimodal} sampling problems, where data is obtained from different random processes. In \S \ref{s:examples} we also present examples where this arises.

\subsection{Contributions}

Besides the general framework described above, our main {contributions} are a series of learning guarantees that relate the amount of training data $m$ to properties of the sampling distributions $\{ \cA_i \}^{m}_{i=1}$. We now present a simplified version of our main result covering the case where $\cA_1 = \cdots = \cA_m = \cA$. The full case is presented in \S \ref{s:learning-guar}.

As noted, a key quantity in our theory is the \textit{variation} of a (nonlinear) set $\bbV$ with respect to a distribution $\cA$ of bounded linear operators in $\cB(\bbX_0,\bbY)$. We define this as the smallest constant $\Phi = \Phi(\bbV; \cA)$ such that  
\be{
\nm{A(v)}^2_{\bbY} \leq \Phi ,\ \forall v \in \bbV,\quad \text{a.s.\ $A \sim \cA$}.
}
See also Definition \ref{def:variation}. As we discuss in Example \ref{ex:comp-sens-coherence}, the variation is effectively a generalization of the notion of \textit{coherence} in classical compressed sensing \cite{candes2011probabilistic}. It also generalizes various generalized notions of coherence, such as the `local coherence', `local coherence in levels' and `block coherence' (see, e.g., \cite{bigot2016analysis,krahmer2013stable}). In addition, as we discuss in Examples \ref{ex:mat-lev-scores} and \ref{ex:christoffel-function}, it is related to so-called \textit{leverage scores} and \textit{Christoffel functions} arising in numerical linear algebra, active learning in machine learning and function regression.

The following result also involves the covering number $\cN(K,d,t)$ of a set $K$ in a pseudometric space (see Definition \ref{def:cov-num}). For convenience, we also write $S(\bbV) = \left \{ v \in \bbV : \nm{v}_{\bbX} = 1\right \}$ and we recall that a set $\bbV$ is a \textit{cone} if $t v \in \bbV$ for all $t \geq 0$ and  $v \in \bbV$.

\thm{
[Simplified main result]
\label{t:main-res-simplified}
Consider the setup of \S \ref{s:setup} with $\alpha = \beta = 1$ and $\cA_i = \cA$, $\forall i$. Let $\bbU$ be a subset of a finite-dimensional subspace of $\bbX_0$ and suppose that $\Delta \bbU : = \bbU - \bbU$ is a cone. Suppose that, for some $0 < \epsilon < 1$,
\be{
\label{main-res-meas-cond-alt-simplified}
m \gtrsim \Phi(S(\Delta \bbU) ; \cA ) \cdot \left [ \left( \int^{1/2}_{0} \sqrt{\log(2 \cN(S(\Delta \bbU) , \nms{\cdot}_{\bbX}, \tau  t  ) ) } \D t  \right)^2 + \log(2/\epsilon) \right ],
}
where $\tau = \sqrt{\frac{\Phi(S(\Delta \bbU) ; \bar{\cA} )}{\Phi(S(\Delta^2 \bbU)) ; \bar{\cA} ) } }$. Let $x \in \bbX_0$, $\theta \geq \nm{x}_{\bbX} $ and $\check{x} = \min \{ 1 , \theta/\nm{\hat{x}}_{\bbX} \} \hat{x}$ for any minimizer $\hat{x}$ of \ef{least-squares-problem} with noisy measurements \ef{noisy-meas}. If $N =\frac{1}{m} \sum^{m}_{i=1} \nm{e_i}^2_{\bbY_i} $, then
\be{
\label{main-res-err-bound-simplified}
\bbE \nm{x - \check{x}}^2_{\bbX} \lesssim  \inf_{u \in \bbU} \nm{x - u}^2_{\bbX} + \theta^2 \epsilon + N.
}
}

This is a simplified version of our main result, Theorem \ref{t:main-res}, which is itself a direct consequence of Corollary \ref{cor:main-res-deltaU-cone}. Besides allowing for nondegeneracy ($\beta \neq \alpha$) instead of isotropy ($\alpha = \beta = 1$) and different distributions $\cA_i$, Theorem \ref{t:main-res} also considers inexact minimizers of \ef{least-squares-problem} and drops the assumption that $\Delta \bbU$ is a cone, at the expense of a more complicated condition in place of \ef{main-res-meas-cond-alt-simplified}.

The measurement condition \ef{main-res-meas-cond-alt-simplified} is the main objective of this paper. It quantifies how many measurements suffice for good generalization, i.e., \ef{main-res-err-bound-simplified}, in terms of the product of the variation of $S(\Delta \bbU)$ and an entropy integral over the same set. The former describes how the measurements interact with the model class $\bbU$, while the latter is a measure of its intrinsic complexity. Separation into these distinct components is a key facet of our theory, one that will allow us  to analyze various different problems of interest later.

Its presence in \ef{main-res-meas-cond-alt-simplified} means that sampling operators with small variation are desirable. In particular, this means that $\nm{A(v)}_{\bbY}$ should not grow large for elements $v$ of the unit sphere $S(\Delta \bbU)$ of the difference set $\Delta \bbU$. Note that the appearance of $\Delta \bbU$ (instead of, e.g., $\bbU$) is hardly surprising, as a critical component of successful recovery is the ability to distinguish elements of $\bbU$ from their measurements.

The entropy integral in \ef{main-res-meas-cond-alt-simplified} is a measure of the intrinsic complexity of the model class. Theorem \ref{t:main-res-simplified} is consequently very general -- it places minimal assumptions on $\bbU$ -- but it leaves open the problem of estimating the entropy integral. In subsequent results (Corollaries \ref{cor:main-res-deltaU-subspaces-I} and \ref{cor:main-res-deltaU-subspaces-II}) we place additional assumptions on $\bbU$ that lead to more explicit estimates.

An important consequence of how \ef{main-res-meas-cond-alt-simplified} separates into these distinct terms is to provides a mechanism for \textit{active learning}. Since the influence of the measurements in \ef{main-res-meas-cond-alt-simplified} is felt only in the variation, a theoretically-optimal active learning strategy is one that minimizes this quantity (over a class of permitted sampling distributions). We will return to this topic later in the context of generative models.

Having presented our main results in \S \ref{s:learning-guar}, we then describe how this framework unifies, generalizes and, in some cases, improves known results. Specifically:
\begin{enumerate}[(a)]
\item \textit{(Structured) compressed sensing (\S \ref{s:CSapplic}).} We show how classical results on compressed sensing are simple consequences of our main results. Moreover, compressed sensing guarantees with structured sparsity models (e.g., weighted sparsity, joint sparsity, group sparsity, sparsity in levels and so forth) also follow easily from our results.
\item \textit{Compressed sensing with generative models (\S \ref{s:applic-gen-CS}).} We consider the case where $\bbU$ is the range of a Lipschitz map $F : \bbZ \rightarrow \bbX$. This is the general problem encountered in compressed sensing with generative models, which is a popular approach in inverse problems. Recent results in \cite{berk2022coherence,berk2023model} give recovery guarantees for feedforward ReLU neural networks. Here we consider a much broader setting, where $F$ is a Lipschitz map (which need not be a feedforward neural network). Moreover, we also obtain recovery guarantees general types of measurements (which may be vector-valued), while previous works were limited to either random Gaussian or randomly-subsampled unitary measurements. To the best our knowledge, our results are the first to consider general types of measurements and general Lipschitz maps. We also provide a full analysis of the active learning strategy first introduced in \cite{adcock2023cs4ml} and subsequently considered in \cite{berk2023model}, which allows one to significantly enhance the performance of solving inverse problems with generative models.
\end{enumerate}

\subsection{Related work}\label{ss:related}

Our framework is inspired by previous work in compressed sensing, notably the \textit{isotropic vector model} of \cite{candes2011probabilistic}, which was later extended in 
\cite[Chpt.\ 12]{adcock2021compressive} to compressed sensing with \textit{jointly isotropic vectors}. This considers the specific case where $\bbX_0 = \bbC^N$, $\bbY_1 = \cdots = \bbY_m = \bbC$ and $\bbU$ is the set of $s$-sparse vectors, i.e., the target object is an (approximately) sparse vector and the sampling operators are linear functionals. Note that isotropy corresponds $\alpha = \beta = 1$ in \ef{nondegeneracy}. We relax this to allow $\alpha \neq \beta$. Within the compressed sensing literature, there are a number of works that allow for non-scalar valued measurements. See \cite{bigot2016analysis,boyer2019compressed} for an instance of vector-valued measurements (`block sampling') and \cite{traonmilin2017compressed} as well as \cite{adcock2024efficient,dexter2019mixed} for Hilbert-valued measurements.

Recovery guarantees in compressed sensing are generally derived for specific model classes, such as sparse vectors or various generalizations (e.g., joint sparse vectors, group sparse vectors, sparse in levels vectors, and so forth \cite{adcock2017breaking,baraniuk2010model-based,bourrier2014fundamental,davenport2012introduction,duarte2011structured,traonmilin2018stable}. Guarantees for general model classes are usually only presented in the case of (sub)Gaussian random measurements (see e.g., \cite{baraniuk2010model-based,dirksen2016dimensionality}). These, while mathematically elegant, are typically not useful in practice. Our framework provides a unified set of recovery guarantees for very general types of measurements. It contains \textit{subsampled unitary matrices} (a well-known measurement modality in compressed sensing, with practical relevance -- see Example \ref{ex:comp-sens-sub-unitary}) as a special case, but also many others, including non-scalar valued measurements.

The proofs of our main results use some techniques from classical compressed sensing (see, e.g., \cite[Chpt.\ 12]{foucart2013mathematical} or \cite[Chpt.\ 13]{adcock2021compressive}). In particular, they rely on Dudley's inequality, Maurey's lemma and a version of Talagrand's theorem. Our work involves the significant generalization of these arguments to, firstly, much broader classes of sampling problems (i.e., not just linear functionals of finite vectors), and secondly, to arbitrary model classes, rather than classes of (structured) sparse vectors. Our results also broaden and strengthen recent results in the active learning context found in \cite{adcock2023cs4ml} and \cite{eigel2022convergence}. In particular, \cite{adcock2023cs4ml} assumes a union-of-subspaces model for $\bbU$ and then uses matrix Chernoff-type estimates. This is similar (although less general in terms of the type of measurements allowed) to our condition (c) in Corollary \ref{cor:main-res-deltaU-subspaces-I}. The authors of \cite{eigel2022convergence} make very few assumptions on $\bbU$, then use Hoeffding's inequality and covering number arguments. As noted in \cite[\S A.3]{adcock2023cs4ml} the trade-off for this high level of generality is weaker theoretical guarantees.

Compressed sensing with generative models was introduced in \cite{bora2017compressed}, and has proved very effective in image reconstruction tasks such as Magnetic Resonance Imaging (MRI) (see \cite{berk2022coherence,jalal2021robust} and references therein). Initial learning guarantees for generative models involved (sub)Gaussian random measurements \cite{bora2017compressed,chen2023unified,jalal2021instance}. Guarantees for subsampled unitary matrices were established in \cite{berk2022coherence,berk2023model} for feedforward ReLU generative models. As noted above, our work significantly extends these existing results to general measurements and general Lipschitz maps.
\rem{
This paper is an extension of an earlier conference publication by the authors \cite{adcock2024unified}. In this paper, we broaden and improve this work in the following key ways. 
\begin{enumerate}[(i)]
\item Our main result (Theorem \ref{t:main-res}) is substantially more general, in that it allows for almost arbitrary $\bbU$. The main results in \cite{adcock2024unified} assumes a \textit{union-of-subspaces} model for $\bbU$ (see Example \ref{ex:UofS}), which is far more restrictive. They are now presented as Corollaries \ref{cor:main-res-deltaU-subspaces-I} and \ref{cor:main-res-deltaU-subspaces-II} of our main result.
\item In \S \ref{s:applic-gen-CS}, we present the first results for model classes defined by Lipschitz maps, with application to generative models, for general measurements. Only ReLU neural networks were considered in  \cite{adcock2024unified}, and the techniques involved are completely different. The analysis of this case relies critically on the generalization (i).

\item We generalize the setup of \cite{adcock2024unified} to seamlessly allow for a mixture of deterministic \textit{and} random measurements. In particular, we extend the definition of variation considered in \cite{adcock2024unified}. See Definition \ref{def:variation-collection}. As we discuss in Example \ref{ex:half-half-multilevel}, this extension is important in applications. 
\item Previous results in \cite{adcock2024unified} for \textit{subsampled unitary matrices} (an important setting in practice, see Example \ref{ex:comp-sens-sub-unitary}) considered sampling with replacement, which has some practical disadvantages. We show that sampling without replacement via Bernoulli selectors can also be analyzed within our framework (Example \ref{ex:bernoulli-model}). The analysis of this setting also relies crucially on the generalization introduced in (iii).
\end{enumerate}
}

\subsection{Outline}

The remainder of this paper proceeds as follows. First, in \S \ref{s:examples} we present a series of examples that showcase the generality of the framework introduced in \S \ref{s:setup}. Next, we introduce some key concepts in \S \ref{s:key-concepts} followed by our main results in \S \ref{s:learning-guar}. We then discuss its application to (structured) compressed sensing in \S \ref{s:CSapplic} and compressed sensing with generative models in \S \ref{s:applic-gen-CS}. Finally, in \S \ref{ss:cov-num-prop}--\ref{s:proofs-end} we give the proofs of our main theorems.

\section{Examples}\label{s:examples}

We now present a series of examples.

\subsection{Sampling problems}

We first consider different sampling problems that can be cast into this unified framework.

\examp{[Regression from i.i.d.\ samples]
\label{ex:function-regression}
Let $D \subseteq \bbR^d$ be a domain with a measure $\rho$ and consider learning an unknown function $f  \in L^2_{\rho}(D)$ from data $\{(z_i,f(z_i))\}^{m}_{i=1}$, where $z_i \sim_{\mathrm{i.i.d.}} \mu$ for some probability measure $\mu$ on $D$. To cast this problem in our framework, we make the (mild) assumption that $\mu$ is absolutely continuous and $\nu : = \D \mu / \D \rho > 0$ a.e.. Now let $\bbX = L^2_{\rho}(D)$, $\bbX_0 = C(\overline{D})$, $\bbY_i = \bbY = \bbR$, $\forall i$, with the Euclidean inner product and $\cA_i = \cA$, $\forall i$, be defined by $A \sim \cA$ if $A(f) = f(z) / \sqrt{\nu(z)}$ for $z \sim \mu$. Note that nondegeneracy \ef{nondegeneracy} holds with $\alpha = \beta = 1$ in this case. Now, given an approximation space $\bbU \subseteq C(\overline{D})$, the least-squares problem \ef{least-squares-problem} becomes the (nonlinear) weighted-least squares fit
\be{
\label{wLS-regression}
\hat{f} \in \argmin{u \in \bbU} \frac1m \sum^{m}_{i=1} \frac{1}{\nu(z_i)} | f(z_i) + \sqrt{\nu(z_i)} e_i - u(z_i) |^2.
}
Note that it is common to set $\mu = \rho$ in such problems, in which case $\nu = 1$ and \ef{wLS-regression} is an unweighted least-squares fit. However, this more general setup allows one to consider the active learning setting, where the sampling measure $\mu$ is chosen judiciously in term of $\bbU$ to improve the learning performance of $\hat{f}$. See Example \ref{ex:christoffel-function} for further discussion.
}

\examp{[Matrix sketching for large least-squares problems]
\label{ex:matrix-sketching}
Let $X \in \bbC^{N \times n}$, $N \geq n$ and $y \in \bbC^N$. In many applications, it is infeasible (due to computational constraints) to find a solution to the `full' least-squares problem
\bes{
w \in \argmin{z \in \bbC^n} \nm{X z - x}_{\ell^2}.
}
Therefore, in matrix sketching \cite{malik2026fast,woodruff2014sketching} one aims to find a \textit{sketching matrix} $S \in \bbC^{m \times N}$ (a matrix with one nonzero per row) such that a minimizer
\be{
\label{sketching-min}
\hat{w} \in \argmin{z \in \bbC^n} \nm{S X z - S x}_{\ell^2}
}
satisfies
\be{
\label{matrix-sketching-goal}
\nm{X \hat{w} - y}^2_{\ell^2} \lesssim \nm{X w - x}^2_{\ell^2} .
}
A particularly effective way to do this involves constructing a random sketch. Let $\pi = \{\pi_1,\ldots,\pi_N\}$  be a discrete probability distribution on $[N] : = \{1,\ldots,N\}$, i.e., $0 < \pi_i \leq 1$ and $\sum^{N}_{i=1} \pi_i = 1$. Then we draw $m$ integers $j_1,\ldots,j_m \sim_{\mathrm{i.i.d.}} \pi$ and set $S_{ij} = 1/\sqrt{\pi_{j_i}}$ if $j = j_i$ and $S_{ij} = 0$ otherwise, so that $S X \in \bbC^{m \times n}$ consists of $m$ rows of $X$ scaled by the probabilities $1/\sqrt{\pi_i}$. 

To cast this problem into the above framework we let $\bbX = \bbX_0 = \bbC^N$ and $\bbY = \bbC$ both equipped with the Euclidean norm. Note that bounded linear operators $\bbX_0 \rightarrow \bbY$ are equivalent to column vectors $a \in \bbC^N$ (via the relation $x \mapsto a^* x$). Hence we define $\cA_1 = \cdots = \cA_m = \cA$ to be a distribution of vectors in $\bbC^N$ with $a \sim \cA$ if
\bes{
\bbP(a = e_i / \sqrt{\pi_i}) = \pi_i,\quad i = 1,\ldots,N.
}
Observe that nondegeneracy \ef{nondegeneracy} holds with $\alpha = \beta = 1$.
Finally, we consider the model class
\be{
\label{U-sketching}
\bbU = \{ X z : z \in \bbC^n \}.
}
Then we readily see that \ef{least-squares-problem} (with $b_i = (S x)_i$) is equivalent to \ef{sketching-min} in the sense that $\hat{x} = X \hat{w}$ is a solution of \ef{least-squares-problem} if and only if $\hat{w}$ is a solution of \ef{sketching-min}. In particular, $\nm{X \hat{w} - x}^2_{\ell^2} = \nm{\hat{x} - x}^2_{\ell^2}$ is precisely the $\bbX$-norm error of the estimator $\hat{x}$.

\textit{Leverage score} sampling is a near-optimal solution to the matrix sketching problem \cite{woodruff2014sketching}. Here, one sets
\be{
\label{pi-leverage-score}
\pi_i = \tau(X)(i) / n,\quad i = 1,\ldots,N,
}
where
\be{
\label{leverage-score}
\tau(X)(i) = \max_{\substack{z \in \bbC^n \\ X z \neq 0}} \frac{|(X z)_i|^2}{\nm{X z}^2_2},\quad i = 1,\ldots,N,
}
are \textit{leverage scores} of the matrix $X$. In this case, \ef{matrix-sketching-goal} holds with high probability, provided
\be{
\label{matrix-sketch-bound}
m \gtrsim n \cdot \log(2 n / \epsilon).
}
As was shown in \cite[App.\ A]{adcock2024unified}, this well-known bound is straightforward consequence of our general theory. Thus leverage score sampling is a specific case of our unified framework.
}

\examp{[Regression with vector-valued measurements]
\label{ex:func-regress-vector}
Regression problems in various applications call for learning vector- as opposed to scalar-valued functions. These are readily incorporated into this framework by modifying Example \ref{ex:function-regression}. Several such extensions were discussed in \cite{adcock2024unified}. These include: Hilbert-valued functions, where  $f : D \rightarrow \bbV$ and $\bbV$ is a Hilbert space); continuous-in-time sampling, where $f : D \times [0,T] \rightarrow \bbR$ and each measurement takes the form $\{ f(x,t) : 0 \leq t \leq T \}$ for fixed $x \in D$; and gradient-augmented measurements, where $f : D \rightarrow \bbR$ and each measurement has the form $ (f(x) , \nabla f(x)) \in \bbR^{d+1}$. See \cite[Ex.\ 2.1]{adcock2024unified} for further details.
}

In the next four examples, we show how this framework includes as special cases various general sampling models from the compressed sensing literature.

\examp{[Compressed sensing with isotropic vectors]
\label{ex:comp-sens-isotropic}
Classical compressed sensing concerns learning a sparse approximation to a vector $x \in \bbC^N$ from $m$ linear measurements. A well-known model involves sampling with \textit{isotropic vectors} (see \cite{candes2011probabilistic} and \cite[Chpt.\ 11]{adcock2021compressive}). We can cast this model in our framework as follows. Let $\bbX = \bbX_0 = \bbC^N$, $\bbY_i = \bbC$, $\forall i$, both equipped with the Euclidean inner product, and $\cA_i = \cA$, $\forall i$, where $\cA$ is a distribution of vectors in $\bbC^N$ that are \textit{isotropic}, i.e.,
\be{
\label{isotropy-vectors}
\bbE_{a \sim \cA} a a^* = I.
}
Here $I$ is the $N \times N$ identity matrix. Note that \ef{nondegeneracy} holds with $\alpha = \beta = 1$ in this case and the measurements \ef{noisy-meas} have the form
\bes{
b_i = a^*_i x + e_i \in \bbC,\quad i = 1,\ldots,m,
}
where $a_1,\ldots,a_m \sim_{\mathrm{i.i.d.}} \cA$. In matrix-vector notation, we can write this as $b = A x + e$, where 
\be{
\label{cs-meas-matrix}
A = \frac{1}{\sqrt{m}} \begin{bmatrix} a^*_1 \\ \vdots \\ a^*_m \end{bmatrix} \in \bbC^{m \times N},\ b =  \frac{1}{\sqrt{m}}  \begin{bmatrix} b_1 \\ \vdots \\ b_m \end{bmatrix} \in \bbC^{m} \text{ and } e =  \frac{1}{\sqrt{m}}  \begin{bmatrix} e_1 \\ \vdots \\ e_m \end{bmatrix} \in \bbC^{m},
} 
(the division by $1/\sqrt{m}$ is a convention: due to \ef{isotropy-vectors}, it ensures that $\bbE(A^*A) = I$).

As discussed in \cite{candes2011probabilistic}, this model includes not only the well-known case of subgaussian random matrices, but also many other common sampling models used in signal and image processing applications. It is also a generalization of the concept of random sampling with \textit{bounded orthonormal systems} (see, e.g., \cite[Chpt.\ 12]{foucart2013mathematical}). Moreover, if we slightly relax \ef{isotropy-vectors} to  $\alpha I \preceq \bbE_{a \sim \cA} a a^* \preceq \beta I$,
so that \ef{nondegeneracy} holds with the same values of $\alpha$ and $\beta$, then it also generalizes the bounded Riesz systems model studied in \cite{brugiapaglia2021sparse}.
}

\examp{[Compressed sensing with subsampled unitary matrices]
\label{ex:comp-sens-sub-unitary}
A case of interest within the previous example is the class of \textit{randomly subsampled unitary} matrices (see also \cite[Defn.\ 5.6 and Examp.\ 11.5]{adcock2021compressive}). Let $U \in \bbC^{N \times N}$ be unitary, i.e., $U^* U = I$. For example, $U$ may be the matrix of the Discrete Fourier Transform (DFT) in a Fourier sensing problem. 
Let $u_i = U^* e_i$, where $e_i$ is the $i$th canonical basis vector, and define the (discrete) isotropic distribution of vectors $\cA$ by $a \sim \cA$ if
\bes{
\bbP ( a = \sqrt{N} u_i) = \frac1N,\quad i = 1,\ldots,N.
}
Note that the corresponding measurement matrix \ef{cs-meas-matrix} consists of rows of $U$ -- hence the term `subsampled unitary matrix'. 

This setup can be straightforwardly extended to the case where the rows of $U$ are sampled with different probabilities, termed a \textit{nonuniformly subsampled unitary matrix}. Let $\pi = (\pi_1,\ldots,\pi_N)$ be a discrete probability distribution on $[N]$, i.e., $0 < \pi_i \leq 1$ and $\sum^{N}_{i=1} \pi_i = 1$. Then we now modify the distribution $\cA$ so that $a \sim \cA$ if
\bes{
\bbP(a = u_i/\sqrt{\pi_i} ) = \pi_i,\quad i = 1,\ldots,N.
}
This family is again isotropic.  Note that uniform subsampling corresponds to $\pi_i = 1/N$, $\forall i$. 
Nonuniform subsampling is important in various applications. For instance, in applications that involve Fourier measurements (e.g., MRI, NMR, Helium Atom Scattering and radio interferometry) it is usually desirable to sample rows corresponding to low frequencies more densely than rows corresponding to high frequencies.
}

\examp{[Sampling without replacement]
\label{ex:bernoulli-model}
The previous example considers random sampling of the rows of $U$ \textit{with replacement}, meaning that repeats are possible. This is often undesirable. One way to avoid this is to use \textit{Bernoulli selectors}. Here, the decision of whether or not to include a row of $U$ is based on the outcome of a (biased) coin toss, i.e., a realization of a Bernoulli random variable.
Specifically, let $\pi = (\pi_1,\ldots,\pi_N)$ satisfy $0 < \pi_i \leq 1/m$ and $\sum^{N}_{i=1} \pi_i = 1$. Then define the distribution $\cA_i$ by $a_i \sim \cA_i$ if
\bes{
\bbP \left (a_i = \sqrt{\frac{N}{m \pi_i}} u_i \right ) = m \pi_i,\quad \bbP(a_i = 0) = 1 - m \pi_i.
}
The family $\{ \cA_i \}^{N}_{i=1}$ is nondegenerate and hence this model falls within the general setup. Later, in \S \ref{ss:gen-mod-rec-unitary} and Corollary \ref{cor:gen-mod-unitary-sub-bernoulli}, we will see that this model admits essentially the same learning guarantees as that of Example \ref{ex:comp-sens-sub-unitary}. Observe that the measurement matrix \ef{cs-meas-matrix} is $N \times N$ in this case, where the $i$th row is proportional to $u^*_i$ if the $i$th coin toss returns a heads, and zero otherwise. In particular, it is equivalent to a $q \times N$ matrix, where $q$ is the number of heads. Unlike in the previous model, the number of measurements $q$ is now a random variable with $\bbE(q) = m$.
}

The reader will likely have notice that, the previous example involves different distributions $\cA_i$. This is in , in contrast to Examples \ref{ex:function-regression}--\ref{ex:comp-sens-sub-unitary}, where $\cA_i = \cA$, $\forall i$. We now present several further examples to motivate this aspect of our general framework.

\examp{
[Half-half and multilevel sampling]
\label{ex:half-half-multilevel}
Recall that Example \ref{ex:comp-sens-sub-unitary} arises in various applications involving Fourier measurements. In these applications, it is often desirable to fully sample the low-frequency regime \cite{lustig2007sparse,adcock2021compressive}. The low frequencies of a signal or image contain much of its energy, meaning that any purely random sampling scheme can miss important information with nonzero probability.

A simple way to avoid this problem is a \textit{half-half} sampling scheme \cite{adcock2017breaking,studer2011compressive}. Here the first $m_1$  measurements are set equal to the lowest $m_1$ frequencies (i.e., the first $m_1$ samples are deterministic) and the remaining $m_2 : = m - m_1$ samples are obtained by randomly sampling the remaining $N - m_1$ frequencies from some distribution. This can be formulated within our framework as follows. Let $U \in \bbC^{N \times N}$ be unitary (typically, $U$ is the DFT matrix, but this is not needed for the current discussion) and $u_i = U^* e_i$, as before. Then, for $i = 1,\ldots,m_1$, we define $\cA_i$ by $a \sim \cA_i$ if
\bes{
\bbP(a = \sqrt{m} u_i) = 1,\quad i= 1,\ldots,m_1.
}
In other words, $a \sim \cA_i$ takes value $\sqrt{m} u_i$ with probability one. Now let $\pi = (\pi_{m_1+1},\ldots,\pi_N)$ be a discrete probability distribution on $\{m_1+1,\ldots,N\}$ with $0 < \pi_i \leq 1$ for all $i$. Then we define $\cA_{m_1+1} = \cdots = \cA_{m} = \cA$ with $a \sim \cA$ if
\bes{
\bbP(a = \sqrt{m / (m_2 \pi_i) } u_i ) = \pi_i,\quad i = m_1+1,\ldots,N.
}
We readily see that the family $\{ \cA_i \}^{m}_{i=1}$ is nondegenerate with $\alpha = \beta = 1$. Notice that the first $m_1$ rows of the measurement matrix obtained from this procedure are proportional to the first $m_1$ rows of $U$ and its remaining $m_2 = m - m_1$ rows are randomly sampled according to $\pi$ from the last $N - m_1$ rows of $U$.

Half-half sampling has been used in various imaging applications \cite{adcock2017breaking,lustig2007sparse,chauffert2014variable,romberg2008imaging,studer2011compressive}. Such schemes subdivide the indices $[N]$ into two \textit{levels}: fully sampled and randomly subsampled. In practice, it is often desirable to have further control over the sampling, by subdividing into more than two levels. This is known as \textit{multilevel random subsampling} \cite{adcock2017breaking}. It can be formulated within this framework in a similar way.
}

\examp{[Multimodal data]
Another benefit of allowing for different distributions is that it can model \textit{multimodal} data. Consider the general setup of \S \ref{s:setup}. We now assume that there are $C > 1$ different types of data, with the $c$th type generating $m_c$ samples via a distribution $\cA^{(c)}$, $c = 1,\ldots,C$, of bounded linear operators in $\cB(\bbX_0,\bbY^{(c)})$. Let $m = m_1+\cdots + m_C$ and define $\{ \cA_i \}^{m}_{i=1}$ by
\bes{
\cA_i = \cA^{(c)}\quad \text{if } m_1 + \cdots + m_{c-1} < i \leq m_1+\cdots + m_c.
}
Thus, the first $m_1$ samples are generated by $\cA^{(1)}$, the next $m_2$ samples by $\cA^{(2)}$, and so forth. Notice that nondegeneracy \ef{nondegeneracy} is now equivalent to the condition
\bes{
\alpha \nm{x}^2_{\bbX} \leq \sum^{C}_{c=1} \frac{m_c}{m} \bbE_{A \sim \cA^{(c)}} \nm{A(x)}^2_{\bbY^{(c)}} \leq \beta \nm{x}^2_{\bbX}.
}
Multimodal data was previously considered in \cite{adcock2023cs4ml}, which, as noted in \S \ref{s:introduction} is a special case of this work. Such data arises in various applications, including multi-sensor imaging systems \cite{chun2017compressed} and PINNs for PDEs \cite{han2018solving,raissi2019physics-informed}. This also includes the important case of \textit{parallel} MRI \cite{mcrobbie2006mri}, which is used widely in medical practice. Another application involves an extension of the gradient-augmented learning problem in Example \ref{ex:func-regress-vector} where, due to cost or other constraints, one can only afford to measure gradients in addition to function values at some fraction of the total samples \cite{peng2016polynomial}.
}

\subsection{Model classes}

Our unified framework allows for an arbitrary model class $\bbU$. Later, we impose various assumptions on $\bbU$ to obtain different types of learning guarantees.  We now introduce a number of different model classes that will be the focus of later sections. We stress, though, that other model classes can readily by considered, as Theorem \ref{t:main-res-simplified} is very general. For concrete guarantees, all that is needed are estimates of the relevant entropy integrals. The examples below are cases where such estimates can be readily provided, as we do later.

\examp{
[Subspaces and unions of subspaces]
\label{ex:UofS}
The simplest case is where $\bbU$ is an $n$-dimensional subspace. This was already considered in Example \ref{ex:matrix-sketching} in the context of matrix sketching (see \ef{U-sketching}). Note that for this model, the goal is typically to show good generalization subject to a number of measurements scaling linearly or log-linearly with $n$, i.e., $m = \ord{n}$ or $m = \ord{n \log(n)}$. See \ef{matrix-sketch-bound} and Example \ref{ex:christoffel-function} later. However, many modern learning problems involve nonlinear model classes. A particularly common model is the \textit{union-of-subspaces} model. Here, we assume that $\bbU = \bbU_1 \cup \cdots \bbU_d$, where each $\bbU_i$ is a subspace of dimension at most $n$. In this case, we typically aim to derive measurement conditions scaling log-linearly in $n$ and mildly (e.g., logarithmically) in $d$. We discuss this further in the next example.
}

\examp{
[Sparse and structured sparse models]
\label{ex:sparse-vectors}
Consider the setting of Example \ref{ex:comp-sens-isotropic} and let $1 \leq s \leq N$. Classical compressed sensing considers the set of $s$-sparse vectors
\be{
\label{U-Sigma-s}
\bbU = \Sigma_s = \{ x \in \bbC^N : \text{$x$ is $s$-sparse} \}.
}
Recall that a vector is \textit{$s$-sparse} if it has at most $s$ nonzero entries. This a union-of-subspaces model, since
\be{
\label{Sigma-U-of-S}
\Sigma_{s} = \bigcup_{\substack{S \subseteq [N] \\ |S| = s} } \{ x \in \bbC^N : \supp(x) \subseteq S \}.
}
Here $\supp(x) = \{ i : x_i \neq 0 \}$ is the \textit{support} of $x = (x_i)^{N}_{i=1}$. Notice that this is a union of
\be{
\label{d-val-CS}
d = {N \choose s} \leq \left ( \frac{\E N}{s} \right )^{s}
}
subspaces of dimension $s$. As we discuss in \S \ref{s:CSapplic}, many structured sparse models are also union-of-subspaces models.
}

\examp{
[Compressed sensing with generative models]
\label{ex:generative-models}
As introduced in \cite{bora2017compressed}, compressed sensing with generative models  involves replacing the classical model class $\bbU = \Sigma_s$ in \ef{U-Sigma-s} with the range of a generative neural network that has been trained on data relevant to the learning problem at hand (e.g., MR scans in the case of MRI). In general, let $\bbZ$ be a \textit{latent space} and $F : \bbZ \rightarrow \bbX$ be a known map (typically a neural network, although this is not strictly required). Then one considers the model class
\bes{
\bbU = \mathrm{Ran}(F) = \{ F(z) : z \in \bbZ \}.
}
Often in practice, $\bbZ = \bbR^k$ and $\bbX = \bbR^N$ with the Euclidean inner produc, where $N$ is the number of image pixels/voxels. In this case, we strive for measurement conditions that depend on the latent space dimension $k$ (which measures the complexity of the model class $\bbU$) as opposed to the ambient dimension $N$.
}

\section{Key concepts}\label{s:key-concepts}

We now introduce a number of key concepts needed in order to present our main results.

\subsection{Additional notation and approximate minimizers}\label{ss:additional-stuff}

Let $(\overline{\bbY} , \ip{\cdot}{\cdot}_{\overline{\bbY}})$ be the Hilbert space defined as the direct sum of the $\bbY_i$, i.e., $\overline{\bbY} = \bbY_1 \oplus \cdots \oplus \bbY_m$
and $\bar{\cA}$ be the distribution of bounded linear operators in $\cB(\bbX_0,\overline{\bbY})$ induced by the collection $\{ \cA_i \}^{m}_{i=1}$. In other words, $\bar{A} \sim \bar{\cA}$ if  $\bar{A}(x) = (A_1(x),\ldots,A_m(x))$,
where the $A_i$ are independent with $A_i \sim \cA_i$ for each $i$. Observe now that nondegeneracy \ef{nondegeneracy} is equivalent to
\be{
\label{nondegeneracy-equiv}
\alpha \nm{x}^2_{\bbX} \leq \frac1m \bbE_{\bar{A} \sim \bar{\cA}} \nm{\bar{A}(x)}^2_{\overline{\bbY}} \leq \beta \nm{x}^2_{\bbX},\quad \forall x \in \bbX_0,
}
and the least-squares problem \ef{least-squares-problem} is equivalent to
\be{
\label{least-squares-problems-equiv}
\hat{x} \in \argmin{u \in \bbU} \frac1m \nm{\bar{b} - \bar{A}(u)}^2_{\overline{\bbY}},
}
where $\bar{b} = (b_1,\ldots,b_m) \in \overline{\bbY}$. For convenience, we also write $\bar{e} = (e_1,\ldots,e_m) \in \overline{\bbY}$.

In our main results, we consider approximate minimizers of \ef{least-squares-problem} or, equivalently, \ef{least-squares-problems-equiv}. We now define the following.
\defn{
[$\gamma$- and $(\gamma,\theta)$-minimizers]
\label{def:gamma-theta-min}
Given $\gamma \geq 0$ we say that $\hat{x} \in \bbU$ is a \textit{$\gamma$-minimizer} of \ef{least-squares-problem} if
\bes{
\frac1m \sum^m_{i =1} \nm{b_i - A_i(\hat{x})}^2_{\bbY_i} \leq \min_{u \in \bbU} \frac1m \sum^m_{i =1} \nm{b_i - A_i(u)}^2_{\bbY_i} + \gamma.
}
Next, let $\theta \geq 0$. Then $\check{x} \in \bbX_0$ is a \textit{$(\gamma,\theta)$-minimizer} of \ef{least-squares-problem} if $\check{x} = \min \{ 1  , \theta / \nm{\hat{x}}_{\bbX} \} \hat{x}$
for some $\gamma$-minimizer $\hat{x}$ of \ef{least-squares-problem}.
}
We introduce the latter concept for technical reasons. As shown in our proofs, it may not be possible in general to bound the expected error $\bbE \nm{x - \hat{x}}^2_{\bbX}$. However, we are able to bound $\bbE \nm{x - \check{x}}^2_{\bbX}$ whenever $\theta$ is chosen suitably large. See Remark \ref{rem:on-error-bounds} for further discussion.

\subsection{Shifted and difference sets, cones and projections on the unit sphere}

Given a set $\bbV \subseteq \bbX_0$ and an $x^* \in \bbX_0$, we write 
\bes{
\bbV^* = \bbV_{x^*} : = \bbV - \{ x^* \} = \{ v - x^* : v \in \bbV \} \subseteq \bbX_0
}
for the shifted set. We write
\bes{
\Delta \bbV = \bbV - \bbV = \{ v_1 - v_2 : v_1,v_2 \in \bbV \} \subseteq \bbX_0
}
for the difference set and
\bes{
\Delta^2 \bbV = \Delta(\Delta \bbV) = \{ v_1 - v_2 : v_1,v_2 \in \Delta \bbV \}.
}
We also define the projection of $\bbV$ onto the unit sphere as
\be{
\label{S-def}
S(\bbV) = \left \{ \frac{v}{\nm{v}_{\bbX}} : v \in \bbV \backslash \{ 0 \} \right \}.
}
We say that $\bbV$ is a \textit{cone} if $t v \in \bbV$ whenever $t \geq 0$ and $v \in \bbV$. Notice that in this case the set \ef{S-def} is given by
\bes{
S(\bbV) = \left \{ v \in \bbV : \nm{v}_{\bbX} = 1 \right \}
}
i.e., it is intersection of $\bbV$ with the unit sphere. 

Observe that nondegeneracy \ef{nondegeneracy} implies that the quantity
\be{
\label{equiv-X-norm}
\tnm{x}_{\bbX} : = \sqrt{\frac1m \sum^{m}_{i=1} \bbE_{A_i \sim \cA_i} \nm{A_i(x)}^2_{\bbY_i} } = \sqrt{\frac1m \bbE_{\bar{A} \sim \bar{\cA}} \nm{\bar{A}(x)}^2_{\overline{\bbY}}},\quad x \in \bbX_0,
}
defines a norm on $\bbX_0$ that is equivalent to $\nms{\cdot}_{\bbX}$. Specifically,
\bes{
\sqrt{\alpha} \nm{x}_{\bbX} \leq \tnm{x}_{\bbX} \leq \sqrt{\beta} \nm{x}_{\bbX},\quad \forall x \in \bbX_0.
}
Given this norm and a set $\bbV \subseteq \bbX_0$, we define the analogue of \ef{S-def} by
\bes{
\widetilde{S}(\bbU) = \{ u / \tnm{u}_{\bbX} : u \in \bbU \backslash \{0\} \}.
}

\subsection{Covering numbers}

Our main results make use of covering numbers.
\defn{
[Covering number]
\label{def:cov-num}
Let $(\bbM,d)$ be a pseudometric space, $K \subseteq \bbM$ and $t > 0$. A \textit{$t$-covering of $K$} is a subset $C \subseteq K$ such that
\bes{
K \subseteq \bigcup_{x \in C} B(x,t),\quad \text{where } B(x,t) = \{ y \in \bbM : d(x,y) \leq t \}.
}
The \textit{covering number} $\cN(K,d,t)$ is the minimum cardinality of a $t$-covering $C$ of $K$.
}

Note that  $d$ is a \textit{pseudometric} on $\bbM$ if it satisfies all the properties of a metric, except positivity (i.e., $d(x,y)$ may equal zero for $x \neq y$).

\subsection{Variation}

We now formally introduce the concept of variation, which is crucial to our analysis.

\defn{[Variation with respect to a distribution]
\label{def:variation}
Let $\bbY$ be a Hilbert space and $\bbV \subseteq \bbX_0$. Consider a distribution $\cA$ of bounded linear operators in $\cB(\bbX_0,\bbY)$. The \textit{variation of $\bbV$ with respect to $\cA$} is the smallest constant $\Phi = \Phi(\bbV ; \cA)$ such that
\be{
\label{Phi-def}
\nm{A(v)}^2_{\bbY} \leq \Phi ,\ \forall v \in \bbV,\quad \text{a.s.\ $A \sim \cA$}.
}
If no such constant exists then we write $\Phi(\bbV ; \cA) = + \infty$
}

\rem{
[Upper bounds for the variation]
\label{rem:var-upper-bounds}
We specify $\Phi$ as the smallest constant such that \ef{Phi-def} holds to ensure that it is well defined. However, in all our results $\Phi$ can be \textit{any} constant such that \ef{Phi-def} holds. We use this property in our proofs at several points.
}

Recall \ef{S-def}.
We will generally consider the variation of the set $S(\bbV)$ rather than $\bbV$ in what follows. Note that in this case, \ef{Phi-def} reads
\bes{
\nm{A(v)}^2_{\bbY} \leq \Phi \nm{v}^2_{\bbX},\quad \forall v \in \bbV,\quad \text{a.s. }A \sim \cA.
}
We now present several examples showing how variation relates to known concepts.

\examp{[Compressed sensing and coherence]
\label{ex:comp-sens-coherence}
Coherence is a well-known concept in compressed sensing (see, e.g., \cite{candes2011probabilistic}). Consider Example \ref{ex:comp-sens-isotropic} with $\bbU = \Sigma_s$ as in \ef{U-Sigma-s}. In this case, $\Phi(S(\bbU) ; \cA)$ is the smallest constant such that
\bes{
| a^* v |^2 \leq \Phi \nm{v}^2_{\ell^2},\quad \forall v \in \Sigma_s,\ \text{a.s.\ $a \sim \cA$}.
}
The \textit{coherence} $\mu = \mu(\cA)$ of the distribution $\cA$ as the smallest constant such that
\be{
\label{cs-coherence}
\nm{a}^2_{\ell^{\infty}} \leq \mu(\cA),\quad \text{a.s.\ $a \sim \cA$}.
}
See \cite{candes2011probabilistic}.
By the Cauchy--Schwarz inequality, we have $| a^* v |^2 \leq \mu(\cA) s \nm{v}^2_{\ell^2}$ for all $v \in \Sigma_s$. Hence
\be{
\label{comp-sens-variation}
\Phi(S(\Sigma_s) ; \cA) \leq \mu(\cA) s.
}
Thus, the variation is bounded by the coherence $\mu(\cA)$ multiplied by the sparsity $s$. Hence, by Remark \ref{rem:var-upper-bounds}, all subsequent bounds for $\Sigma_s$ can be formulated in terms of $\mu(\cA) s$.

Now consider Example \ref{ex:comp-sens-sub-unitary}, in which case $\cA$ is a distribution of scaled rows of a unitary matrix $U = (u_{ij})^{N}_{i,j=1}$. In this case, we have
\be{
\label{U-coherence}
\mu(\cA) = N \max_{ij} |u_{ij}|^2 = : \mu(U),
}
where $\mu(U)$ is known as the \textit{coherence} of the matrix $U$ \cite{donoho2003optimally,donoho2001uncertainty}. Note that $1 \leq \mu(U) \leq N$. A matrix is said to be \textit{incoherent} if $\mu(U) \approx 1$. Equivalently, the variation is $\approx s$, i.e., close to the complexity of the model class.

In general, coherence directly determines the learning guarantee for sparse vector recovery. A well-known measurement condition (see, e.g., \cite[Cor.\ 13.15]{adcock2021compressive}) takes the form
\be{
\label{m-CS-isotropic}
m \gtrsim \mu(\cA) \cdot s \cdot \left ( \log^2(s) \log(N) + \log(\epsilon^{-1}) \right ).
}
Hence the number of measurements that suffices for good generalization scales linearly in $s$, up to the coherence $\mu(\cA)$ and a polylogarithmic factor. 
In \S \ref{ss:classical-CS-theory} we show that this bound is a straightforward corollary of our main theorems and \ef{comp-sens-variation} when applied to this problem. Hence our framework generalizes classical compressed sensing with isotropic vectors.
}

\examp{
[Matrix leverage scores]
\label{ex:mat-lev-scores}
Consider Example \ref{ex:matrix-sketching} with the model class $\bbU$ given by \ef{U-sketching}. As shown in \cite[App.\ A]{adcock2024unified}, the variation in this case is related to the \textit{leverage scores} \ef{leverage-score} of the matrix $X$. Specifically,
\bes{
\Phi(S(\bbU) ; \bar{\cA}) = \Phi(S(\bbU) ; \cA ) = \max_{i=1,\ldots,N} \left \{ \frac{\tau(X)(i) }{\pi_i} \right \},
}
In particular, minimizing the variation amongst all probability distributions $\pi$ leads precisely to \textit{leverage score sampling} \ef{pi-leverage-score}. See \cite{adcock2024unified} for additional discussion.
}

\examp{
[Christoffel functions and Christoffel sampling]
\label{ex:christoffel-function}
Consider Example \ref{ex:function-regression}. The \textit{Christoffel function} of $\bbU$ is defined by
\bes{
K(\bbU)(z) = \sup \left \{ \frac{|u(z)|^2}{\nm{u}^2_{L^2_{\rho}(D)} } : u \in \bbU,\ u \neq 0 \right \}.
}
It is related to the variation of $\bbU$ in the following way:
\be{
\label{variation-Christoffel}
\Phi(S(\bbU) ; \bar{\cA} ) = \Phi(S(\bbU) ; \cA ) = \esssup_{z \sim \rho} \left \{ \frac{K(\bbU)(z)}{\nu(z)} \right \}.
}
Christoffel functions have recently been considered in depth in function regression problems \cite{adcock2023cs4ml,cohen2017optimal,adcock2025optimal2,avron2017random,chen2016statistical,chen2019active,derezinski2018leveraged,erdelyi2020fourier,gajjar2023active}. Much like in Examples \ref{ex:mat-lev-scores}, one can exploit \ef{variation-Christoffel} in an active learning context by optimizing the sampling measure $\mu$. Specifically, \ef{variation-Christoffel} is minimized amongst all admissible probability measures $\mu$ by choosing
\bes{
\D \mu(z) = \frac{K(\bbU)(z)}{\int_{D} K(\bbU)(z) \D \rho(z)} \D \rho(z).
}
This is known as Christoffel sampling. In particular, if $\bbU$ is an $n$-dimensional subspace, this approach leads to a sample complexity bound that is near-optimal, namely, log-linear in $n$. See \cite[Sec.\ 4]{adcock2024unified} for an in-depth discussion of this approach.
}

We conclude this section with several additional definitions that will be used later.

\defn{
[Constant distribution]
A distribution $\cD$ is \textit{constant} if $X \sim \cD$ takes only a single value $x$ with probability one, i.e., $\bbP(X = x) = 1$. Otherwise it is \textit{nonconstant}. 
}

\defn{
[Variation with respect to a collection]
\label{def:variation-collection}
Let $\bar{\cA}$ be as in \S \ref{ss:additional-stuff}, where each $\cA_i$ is a distribution of bounded linear operators in $\cB(\bbX_0,\bbY_i)$, and
\be{
\label{index-set-I}
\cI = \{ i:  \text{$\cA_i$ is nonconstant} \}\subseteq [m]. 
}
We define the \textit{variation of $\bbV$ with respect to $\bar{\cA}$} as 
\be{
\label{variation-collection}
\Phi(\bbV ; \bar{\cA}) = \max_{i \in \cI} \Phi(\bbV ; \cA_i),
}
and write $\Phi(\bbV ; \bar{\cA}) = + \infty$ if no such constant exists.
}

The motivation for these definitions is as follows. Often in practice, one has some measurements that should always be included. For example, in a function regression problem, it may be desirable to sample the function at certain locations, based on known behaviour.  Similarly, in a Fourier sampling problem, as discussed in Example \ref{ex:half-half-multilevel}, one may wish to fully sample all low frequencies up to some given maximum.
Mathematically, as demonstrated in Example \ref{ex:half-half-multilevel}, we can incorporate a deterministic measurement by defining a distribution that takes a single value with probability one. Since these measurements are deterministic, it is unnecessary to include them in \ef{variation-collection}, since, as we will see, variation pertains to the number of random measurements needed for recovery. Later, when we consider sampling without replacement (Example \ref{ex:bernoulli-model}), we will see that it is, in fact, vital that the variation be defined as a maximum  over the nonconstant distributions only. See Corollary \ref{cor:gen-mod-unitary-sub-bernoulli} and Remark \ref{rem:why-nonconstant}.

\section{Main results}\label{s:learning-guar}

We now present our main results. We commence with a general result, Theorem \ref{t:main-res}, followed by a sequence of corollaries that deal with progressively more specialized cases.

\subsection{General result}

We first require one additional quantity. Let $\{ \cA_i \}^{m}_{i=1}$ be a nondegenerate family of distributions and $\widetilde{m}$ be the number of nonconstant distributions, i.e., $\widetilde{m} = |\cI|$ where $\cI$ is as in \ef{index-set-I}. Fix a realization $\bar{A} = (A_1,\ldots,A_m)$ from $\bar{\cA} = \{ \cA_i \}^{m}_{i=1}$. Given $q \geq 1$, we define the seminorm $\nms{\cdot}_{q,\bar{A}}$ on $\bbX_0$ by
\be{
\label{discrete-norm-dudley-def}
\nm{x}_{q,\bar{A}} = \left ( \frac{1}{\widetilde{m}} \sum_{i \in \cI} \nm{A_i(x)}^{2q}_{\bbY_i} \right )^{\frac{1}{2q}},\quad \forall x \in \bbX_0.
}

\thm{
[General result]
\label{t:main-res}
Consider the setup of \S \ref{s:setup} and let $\bbU$ be a subset of a finite-dimensional subspace of $\bbX_0$. Fix $x^* \in \bbX_0$, let $\bbU^{*} = \bbU - \{ x^* \}$ and suppose that, for some $0 < \epsilon < 1$ and $q \geq 1$,
\be{
\label{main-res-meas-cond}
m \gtrsim \upsilon^{2+\frac2q} \cdot \left [  \left( \esssup_{\bar{A} \sim \bar{\cA}} \int^{1/2}_{0} \sqrt{\log(2 \cN(\widetilde{S}(\bbU^*) , \nms{\cdot}_{q,\bar{A}} , \upsilon t  ) ) } \D t  \right)^2 + \log(2/\epsilon) \right ],
}
where $\upsilon = \sqrt{\Phi(S(\bbU^*) ; \bar{\cA})  / \alpha}$.
Let $x \in \bbX_0$, $\theta \geq \nm{x}_{\bbX} $ and $\gamma \geq 0$. Then
\be{
\label{main-res-err-bd}
\bbE \nm{x - \check{x}}^2_{\bbX} \lesssim \frac{\beta}{\alpha}  \left (  \nm{x - x^*}^2_{\bbX} + \inf_{u \in \bbU} \nm{x - u}^2_{\bbX} \right )+ \theta^2 \epsilon + \frac{\gamma^2}{\alpha} + \frac{\nm{e}^2_{\overline{\bbY}}}{\alpha}
}
for any $(\gamma,\theta)$-minimizer $\check{x}$ of \ef{least-squares-problem} with noisy measurements \ef{noisy-meas}. Moreover, the same result holds with \ef{main-res-meas-cond} replaced by
\be{
\label{main-res-meas-cond-alt}
m \gtrsim \alpha^{-1} \cdot \Phi(S(\bbU^*) ; \bar{\cA}) \cdot \left [ \left( \int^{1/2}_{0} \sqrt{\log(2 \cN(\widetilde{S}(\bbU^*) , \tnm{\cdot}_{\bbX}, \tau  t  ) ) } \D t  \right)^2 + \log(2/\epsilon) \right ],
}
where $\tau = \sqrt{ \Phi(S(\bbU^*) ; \bar{\cA} ) /\Phi(S(\Delta \widetilde{S}(\bbU^*) ) ; \bar{\cA} )}$. 
}

This result establishes a generalization bound \ef{main-res-err-bd} whenever the measurement condition \ef{main-res-meas-cond} or \ef{main-res-meas-cond-alt} holds. These conditions relate the number of measurements $m$ to the variation $\Phi(S(\bbU^*) ; \bar{\cA})$ of the projected shifted set $S(\bbU^*)$ and entropy integrals of $\widetilde{S}(\bbU^*)$ with respect to certain norms. While \ef{main-res-meas-cond} may be difficult to interpret on its own, its generality is needed to derive all subsequent results. In particular, \ef{main-res-meas-cond-alt} follows directly from \ef{main-res-meas-cond} by bounding the empirical seminorm \ef{discrete-norm-dudley-def} by $\tnm{\cdot}_{\bbX}$, using standard properties of covering numbers and then by selecting a suitable value of $q$ (see the proof for details).  

Theorem \ref{t:main-res} demonstrates how the sample complexity is determined by the product of the variation (which is influenced by the sampling operators) and entropy integrals (which measure the complexity of the model class). This splitting is a key aspect of our theory and critical for subsequent examples. Overall, Theorem \ref{t:main-res} is very general, as it not only allows for general types of sampling operators, but also places minimal assumptions on $\bbU$. However, the entropy integrals are not straightforward to interpret at this level of generality. In Corollaries \ref{cor:main-res-deltaU-subspaces-I} and \ref{cor:main-res-deltaU-subspaces-II}, we impose conditions on $\bbU$ that allow these integrals to be estimated, leading to more explicit measurement conditions.

The bound \ef{main-res-err-bd} estimates the expected error up to $\nm{x-x^*}_{\bbX}$, i.e., how well $x$ is approximated by $x^*$, and the best approximation error from the set $\bbU$. The term $\gamma^2/\alpha$ arises because we consider inexact minimizers of \ef{least-squares-problem} and the term $\theta^2 \epsilon$ arises because we consider the truncated estimator $\check{x}$ (see  Definition \ref{def:gamma-theta-min} and Remark \ref{rem:on-error-bounds}). The final term in \ef{main-res-err-bd} involves the noise. Notice that the nondegeneracy constants $\alpha,\beta$ influence the bound. Unsurprisingly, the bound is worse if $\alpha$ is small and/or $\beta$ is large.

Theorem \ref{t:main-res} provides a guarantee that is local to the point $x^*$, which may be chosen arbitrarily. The measurement conditions involve the variation and entropy integrals over the shifted set $\bbU^*$, and the error bound \ef{main-res-err-bd} involves the error $\nm{x - x^*}_{\bbX}$.  Besides intrinsic interest, it transpires that this generality, i.e., allowing arbitrary $x^*$, is required in order to establish our main results on generative models in \S \ref{s:applic-gen-CS}. In view of the term $\inf_{u \in \bbU} \nm{x - u}^2_{\bbX}$ that appears in \ef{main-res-err-bd} it would be natural to consider $x^*$ as an element of $\bbU$ that approximates $x$ well, although this is not necessary in Theorem \ref{t:main-res}. However, following this reasoning, in our next result, we replace $\bbU^*$ by the difference set $\Delta \bbU$, leading to a simplified error bound.

\rem{
[Worst-case bound]
The entropy integral can always be estimated in a crude way, since $\bbU$ is assumed to be a subset of a finite-dimensional space. Suppose that $\bbU^*$ is a subset of a $N$-dimensional subspace $\bbV \subseteq \bbX_0$. Then $\widetilde{S}(\bbU^*)$ is a subset of the unit ball of $\bbV$ with respect to the norm $\tnm{\cdot}_{\bbX}$. It is now a short argument using property (F) of \S \ref{ss:cov-num-prop} and \ef{log-integral-bd} to see that \ef{main-res-meas-cond-alt} is implied  by the condition
\bes{
m \gtrsim \alpha^{-1} \cdot \Phi(S(\bbU^*) ; \bar{\cA}) \cdot \left [ N  \cdot \log(2(1+\tau^{-1} )) + \log(2/\epsilon) \right ].
}
While finite, this condition is typically not useful, as it depends on the ambient dimension $N$ not the intrinsic complexity of the model class. In the case of the sparse model $\Sigma_s$, for instance, the smallest subspace containing $\bbU^*$ is $\bbC^N$ (recall Example \ref{ex:sparse-vectors}). However, the goal in this case is to obtain measurement conditions scaling linearly in $s$, the sparsity, and logarithmically in $N$ (recall \ef{m-CS-isotropic}).
}

\subsection{Bounds involving the difference set $\Delta \bbU$}

We now simplify Theorem \ref{t:main-res} by replacing the shifted set $\bbU^*$ by $\Delta \bbU = \bbU - \bbU$. The next result follows from Theorem \ref{t:main-res} by considering arbitrary $x^* \in \bbU$ and using the fact that $\bbU^* \subseteq \Delta \bbU$.

\cor{
[Bounds involving $\Delta \bbU$]
\label{cor:main-res-deltaU-cone}
Consider the setup of \S \ref{s:setup} and let $\bbU$ be a subset of a finite-dimensional subspace of $\bbX_0$.
Suppose that, for some $0 < \epsilon < 1$ and $q \geq 1$,
\be{
\label{main-res-deltaU-cone-meas-cond-0}
m \gtrsim \upsilon^{2+\frac{2}{q}} \cdot \left [ 
\left( \esssup_{\bar{A} \sim \bar{\cA}} \int^{1/2}_{0} \sqrt{\log(2 \cN(\widetilde{S}(\Delta\bbU) , \nms{\cdot}_{q,\bar{A}} , \upsilon t  ) ) } \D t  \right)^2 
+ \log(2/\epsilon) \right ],
}
where $\upsilon = \sqrt{\Phi(S(\Delta \bbU) ; \bar{\cA})  / \alpha}$.
 Let $x \in \bbX_0$, $\theta \geq \nm{x}_{\bbX} $ and $\gamma \geq 0$. Then
\be{
\label{main-res-deltaU-cone-err-bd}
\bbE \nm{x - \check{x}}^2_{\bbX} \lesssim \frac{\beta}{\alpha}  \inf_{u \in \bbU} \nm{x - u}^2_{\bbX} + \theta^2 \epsilon + \frac{\gamma^2}{\alpha} + \frac{\nm{e}^2_{\overline{\bbY}}}{\alpha},
}
for any $(\gamma,\theta)$-minimizer $\check{x}$ of \ef{least-squares-problem} with noisy measurements \ef{noisy-meas}. Moreover, if $\Delta \bbU$ is a cone, then the same result holds with \ef{main-res-deltaU-cone-meas-cond-0} replaced by either
\be{
\label{main-res-deltaU-cone-meas-cond-1}
m \gtrsim \alpha^{-1} \cdot \Phi(S(\Delta \bbU) ; \bar{\cA}) \cdot \left [ \left( \int^{1/2}_{0} \sqrt{\log(2 \cN(\widetilde{S}(\Delta \bbU) , \tnm{\cdot}_{\bbX} , \tau t  ) ) } \D t  \right)^2 + \log(2/\epsilon) \right ],
}
where $\tau = \sqrt{\frac{\Phi(S(\Delta \bbU) ; \bar{\cA} )}{\Phi(S(\Delta^2 \bbU)) ; \bar{\cA} ) } }$, or
\be{
\label{main-res-deltaU-cone-meas-cond-2}
m \gtrsim \alpha^{-1} \cdot \Phi(S(\Delta^2 \bbU) ; \bar{\cA}) \cdot \left [ \left( \int^{1/2}_{0} \sqrt{\log(2 \cN(\widetilde{S}(\Delta\bbU) , \tnm{\cdot}_{\bbX} , t  ) ) } \D t  \right)^2 + \log(2/\epsilon) \right ].
}
}

This result refines Theorem \ref{t:main-res} by simplifying the error bound \ef{main-res-deltaU-cone-err-bd}. It does so at the expense of a slightly larger measurement condition (\ef{main-res-deltaU-cone-meas-cond-0} versus \ef{main-res-meas-cond}) involving $\Delta \bbU$ instead of $\bbU^*$. This is relevant when one wants to estimate the variation either analytically or numerically. Since one often does not know which $x^*$ ensures a small error in \ef{main-res-meas-cond}, it is desirable to consider $\Delta \bbU$ instead, as this term just depends on the model class $\bbU$. 

\rem{
[$\Delta \bbU$ and $\bbU$ are often similar]
\label{rem:U-DeltaU-similar}
Notice that $\Delta \bbU$ is often structurally similar to $\bbU$ itself. For example, in the sparse model (Example \ref{ex:sparse-vectors}) where $\bbU = \Sigma_s$, we have $\Delta \bbU = \Sigma_{2s}$ and $\Delta^2 \bbU = \Sigma_{4s}$. Similarly, consider the case of generative models (Example \ref{ex:generative-models}) where $\bbU = \mathrm{Ran}(F)$ and $F : \bbR^k \rightarrow \bbR^N$ (for simplicity). Then $\Delta \bbU = \mathrm{Ran}(G)$, where
\bes{
G : \bbR^{2k} \rightarrow \bbR^N,\ (z_1,z_2) \mapsto F(z_1) - F(z_2) = (1,-1)^{\top} (F(z_1) , F(z_2))
} 
and similarly for $\Delta^2 \bbU$. In particular, if $F$ is neural network of a given type, then $G$ is also a neural network of twice the width and, potentially, one additional layer, and its latent dimension is only twice that of $F$.
A key consequence of these observations is that one can typically estimate the variation (analytically or numerically) over $\Delta \bbU$ and $\Delta^2 \bbU$ whenever it is possible to do so for $\bbU$. We will see examples of this later.
}

Corollary \ref{cor:main-res-deltaU-cone} also refines the condition \ef{main-res-meas-cond-alt} in the case where $\Delta \bbU$ is a cone. This is a standard assumption that holds in many cases of interest (this includes all examples we consider in this paper). In particular, \ef{main-res-deltaU-cone-meas-cond-1} simplifies the expression for $\tau$ over that in \ef{main-res-meas-cond-alt} so that it just involves $S(\Delta \bbU)$ and $S(\Delta^2 \bbU)$. Moreover, 	\ef{main-res-deltaU-cone-meas-cond-2} removes $\tau$ altogether, at the expense of considering the variation over the larger, but structurally similar, set $S(\Delta^2 \bbU)$.

\rem{[Difference sets as cones]
Notice that $\Delta \bbU$ is a symmetric set, i.e., $\Delta \bbU = - \Delta \bbU$. Therefore, if $\Delta \bbU$ is a cone, then $t v \in \Delta \bbU$ for all $v \in \Delta \bbU$ and all $t \in \bbR$ and not just $t \geq 0$.
Note that if $\bbU$ is a cone then $\Delta \bbU$ is also a cone. However, the converse may not hold. Indeed, $\bbU = [-1,\infty) \subseteq \bbR$ is not a cone, but its difference set $\Delta \bbU = \bbR$ is a cone.
}

\subsection{The case where $\Delta \bbU$ is contained in a union of subspaces}

The previous results involve entropy integrals. We now refine them by estimating these integrals in the case where  $\Delta \bbU$ is contained in a union of finite-dimensional subspaces. As discussed in Example \ref{ex:sparse-vectors}, this property arises frequently in structured sparse models. In Remark \ref{rem:relu-generative-nn} we discuss how it also arises in the case of ReLU generative neural networks.

\cor{
[$\Delta \bbU$ is contained in a union of subspaces I]
\label{cor:main-res-deltaU-subspaces-I}
Consider the setup of \S \ref{s:setup} and let $\bbU$ be a subset of a finite-dimensional subspace of $\bbX_0$ satisfying
\begin{enumerate}[(i)]
\item $\Delta \bbU$ is a cone, and
\item $\Delta \bbU \subseteq \bbV_1 \cup \cdots \cup \bbV_d = : \bbV$, where each $\bbV_i \subseteq \bbX_0$ is a subspace of dimension at most $n$.
\end{enumerate}
Suppose that, for some $0 < \epsilon < 1$, either
\begin{enumerate}[(a)]
\item $m \gtrsim \alpha^{-1} \cdot \Phi(S(\Delta \bbU) ; \bar\cA) \cdot \left [n  \log \left ( 2 \frac{\min\{\Phi(S(\bbV) ; \bar\cA) , \Phi(S(\Delta^2 \bbU ) ; \bar{\cA}) \}}{\Phi(S(\Delta \bbU) ; \bar{\cA})} \right ) + \log(2 d / \epsilon)  \right ]$,
\item $m \gtrsim \alpha^{-1} \cdot \Phi(S(\Delta^2 \bbU) ; \bar\cA) \cdot \left [ n + \log(2 d / \epsilon)   \right ]$, or
\item $m \gtrsim \alpha^{-1}  \cdot \Phi(S(\bbV) ; \bar\cA) \cdot \left [\log(n) + \log(2 d / \epsilon) \right ]$, where $\bbV$ is as in (ii).
\end{enumerate}
Let $x \in \bbX_0$, $\theta \geq \nm{x}_{\bbX} $ and $\gamma \geq 0$. Then
\be{
\label{main-res-deltaU-cone-err-bd-alt0}
\bbE \nm{x - \check{x}}^2_{\bbX} \lesssim \frac{\beta}{\alpha}  \inf_{u \in \bbU} \nm{x - u}^2_{\bbX} + \theta^2 \epsilon + \frac{\gamma^2}{\alpha} + \frac{\nm{e}^2_{\overline{\bbY}}}{\alpha}
}
for any $(\gamma,\theta)$-minimizer $\check{x}$ of \ef{least-squares-problem} with noisy measurements \ef{noisy-meas}. 
}

This result establishes measurement conditions that involve the variation over $S(\Delta \bbU)$, $S(\Delta^2 \bbU)$ or $S(\bbV)$, the number of subspaces $d$ and their dimensions $n$. Similar to Corollary \ref{cor:main-res-deltaU-cone}, condition (b) improves the log factor over condition (a) at the expense of evaluating the variation over the larger set $S(\Delta^2 \bbU)$. On the other hand, condition (c) improves the linear scaling in $n$ to logarithmic, at the expense of evaluating the variation over $S(\bbV)$.

\rem{
[Why $\Delta \bbU$ and $\Delta^2 \bbU$ not $\bbV$]
\label{rem:relu-generative-nn}
A key aspect of (a) and (b) is that they involve variation over sets related to $\bbU$, not the set $\bbV$. This is important in some applications, such as compressed sensing with generative models (Example \ref{ex:generative-models}). It was shown in \cite[Lem.\ S2.2 \& Rem.\ S2.3]{berk2022coherence} that (i) and (ii) hold with $n = 2k$ and $\log(d) \leq 2k \cdot L$ whenever $F : \bbR^k \rightarrow \bbR^N$ is a feedforward ReLU neural network. Here $L$ is a log factor depending on the width and depth of the network. However, one generally does not have $\Delta \bbU = \bbV$ in this problem.
As we discuss in Remark \ref{rem:compute-coherences}, one can numerically estimate the variation with respect to $\bbU$ (or $\Delta \bbU$, $\Delta^2 \bbU$) in the case of a generative model. By contrast, it is typically infeasible to do this for set $\bbV$. Therefore, it is important to have guarantees such as (a) and (b) that require the variation over $\Delta \bbU$ and $\Delta^2 \bbU$ only.
}

Corollary \ref{cor:main-res-deltaU-subspaces-I} is quite general. However, as we discuss in \S \ref{s:CSapplic}, it yields suboptimal bounds for (structured) sparse models (Example \ref{ex:sparse-vectors}). Specifically, the resulting bounds scale quadratically in $s$, rather than linearly. Fortunately, by making an additional assumption, we can resolve this shortcoming.

\cor{
[$\Delta \bbU$ is contained in a union of subspaces II]
\label{cor:main-res-deltaU-subspaces-II}
Once more, consider the setup of \S \ref{s:setup}. Let $\bbU$ be a subset of a finite-dimensional subspace of $\bbX_0$ satisfying (i) and (ii) of Corollary \ref{cor:main-res-deltaU-subspaces-I}, and also
\begin{enumerate}
\item[(iii)] $\{ u \in \Delta \bbU : \nm{u}_{\bbX} \leq 1 \} \subseteq \mathrm{conv}(\bbW)$, where $\bbW$ is a finite set of size $|\bbW| = M$.
\end{enumerate}
Suppose that, for some $0 < \epsilon < 1$, either
\begin{enumerate}[(a)]
\item $m \gtrsim \alpha^{-1} \cdot \Phi(S(\Delta \bbU) \cup \bbW ; \bar{\cA} ) \cdot L$,
where
\eas{
L = & \log\left (1+  \frac{\min \{ \Phi(S(\Delta^2 \bbU) ; \bar{\cA} ) ,  \Phi(S(\bbV) ; \bar{\cA} ) \} }{\Phi(S(\Delta \bbU) \cup \bbW ; \bar{\cA} )} \right )
\\
&  + \log(2+\Phi(S(\Delta \bbU) \cup \bbW ; \bar{\cA} ) / \alpha ) \cdot \log(2M) \cdot \log^2 \left [ 2 (\log(2d)+n) \right ]+ \log(1/\varepsilon)
}
\item $m \gtrsim \alpha^{-1} \cdot \Phi(S(\Delta^2 \bbU) \cup \bbW ; \bar{\cA} ) \cdot L$, where
\bes{
L = \log(2+\Phi(S(\Delta \bbU) \cup \bbW ; \bar{\cA} ) / \alpha ) \cdot \log(2M) \cdot \log^2 \left [ 2 (\log(2d)+n) \right ] + \log(1/\varepsilon)
} 
\item or $m \gtrsim \alpha^{-1} \cdot \Phi(S(\bbV) \cup \bbW ; \bar{\cA} ) \cdot L$, where $\bbV$ is as in (ii) of Corollary \ref{cor:main-res-deltaU-subspaces-I} and
\bes{
L = \log(2+\Phi(S(\bbV) \cup \bbW ; \bar{\cA} ) / \alpha ) \cdot \log(2M) \cdot \log^2 \left [ 2 (\log(2d)+n) \right ] + \log(1/\varepsilon).
}
\end{enumerate}
Let $x \in \bbX_0$, $\theta \geq \nm{x}_{\bbX} $ and $\gamma \geq 0$. Then
\be{
\label{main-res-deltaU-cone-err-bd-alt}
\bbE \nm{x - \check{x}}^2_{\bbX} \lesssim \frac{\beta}{\alpha}  \inf_{u \in \bbU} \nm{x - u}^2_{\bbX} + \theta^2 \epsilon + \frac{\gamma^2}{\alpha} + \frac{\nm{e}^2_{\overline{\bbY}}}{\alpha}
}
for any $(\gamma,\theta)$-minimizer $\check{x}$ of \ef{least-squares-problem} with noisy measurements \ef{noisy-meas}. 
}

The main difference between this result and Corollary \ref{cor:main-res-deltaU-subspaces-I} is (iii), which states that the unit ball of $\Delta \bbU$ should be contained in the convex hull of a set $\bbW$ that is not too large (since $M$ enters logarithmically in the measurement condition) and has small variation (since the variation is taken over a superset of $\bbW$). Assumption (iii) allows the dependence of the measurement condition on $d$ and $n$ to be reduced from essentially $\log(2d) + n$ in Corollary \ref{cor:main-res-deltaU-subspaces-I} to $\log^2(\log(2d) + n)$. We will exploit this fact critically in \S \ref{s:CSapplic}. We remark in passing that assumption (iii) always holds for some set $\bbW$ whenever assumptions (i) and (ii) of Corollary \ref{cor:main-res-deltaU-subspaces-I} hold. However, the resulting $\bbW$ may not lead to a good bound in the measurement condition. See Remark \ref{rem:ass-iii-holds}. As we discuss next, though, for structured sparse models we can indeed derive a suitable $\bbW$ that leads to a significantly better measurement condition than those implied by Corollary \ref{cor:main-res-deltaU-subspaces-I}.

\section{Application to (structured) compressed sensing}\label{s:CSapplic}

We now illustrate how our main results can be used to establish known results for classical compressed sensing, as well as various types of `structured' compressed sensing

\subsection{Classical compressed sensing with isotropic vectors}
\label{ss:classical-CS-theory}

Consider Example \ref{ex:comp-sens-isotropic}, with $\bbU = \Sigma_s$ as in \ef{U-Sigma-s}. Note that $\cA_i = \cA$, $\forall i$, in this case, where $\cA$ is as in Example \ref{ex:comp-sens-coherence}, and recall that $\alpha = \beta = 1$ and that $\Delta \bbU = \bbU - \bbU = \Sigma_{2s}$. Assumption (i) of Corollary \ref{cor:main-res-deltaU-subspaces-I} holds, since $t x$ is $s$-sparse for any $t \geq 0$ (in fact, $t \in \bbR$) whenever $x$ is $s$-sparse, and assumption (ii) also holds, due to \ef{Sigma-U-of-S}. Example \ref{ex:comp-sens-coherence} derives an expression \ef{comp-sens-variation}  for the variation in terms of $s$ and $\mu(\cA)$. Using this and \ef{d-val-CS}, we deduce that the measurement condition in Corollary \ref{cor:main-res-deltaU-subspaces-I}(a) reduces to
\bes{
m \gtrsim \mu(\cA) \cdot s \cdot \left ( s \log(\E N / s) + \log(\epsilon^{-1}) \right )
}
(one can readily verify that the conditions (b) and (c) give the same result, up to constants). This bound is suboptimal, since it scales quadratically in $s$. Fortunately, this can be overcome by using Corollary \ref{cor:main-res-deltaU-subspaces-II}. It is well known that (see, e.g., \cite[proof of Lem.\ 12.37]{foucart2013mathematical})
\be{
\label{conv-hull-sparsity}
\{ x \in \Sigma_{s} : \nm{x}_2 \leq 1 \} \subseteq \mathrm{conv} \left ( \left \{ \pm \sqrt{2 s} e_i , \pm \sqrt{2 s} \I e_i : i = 1,\ldots,N \right \} \right ) = : \mathrm{conv}(\bbW).
}
This set has size $M = | \bbW | = 4 N$. In order to apply Corollary \ref{cor:main-res-deltaU-subspaces-II} we estimate the variation
\bes{
\Phi(S(\Delta \bbU) \cup \bbW ; \bar{\cA}) = \max \left \{ \Phi(S(\Delta \bbU) ; \bar{\cA}) , \Phi(\bbW ; \bar{\cA}) \right \}.
}
The first term is bounded by $2 s \mu(\cA)$ (recall \ef{comp-sens-variation}). For the latter, we observe that, for any $x \in \bbW$, we have, for some $1 \leq i \leq N$,
\bes{
| a^* x |^2 = 2 s | a^* e_i |^2 \leq 2 s \mu(\cA),\quad \text{a.s. $a \sim \cA$}. 
}
Hence $\Phi(\bbW ; \bar{\cA}) \leq 2 s \mu(\cA)$ as well. Using this and \ef{d-val-CS} we see that condition (a) of Corollary \ref{cor:main-res-deltaU-subspaces-II} (the same applies for conditions (b) and (c)) reduces to
\be{
\label{goodr-sunglasses}
m \gtrsim \mu(\cA) \cdot s \cdot \left ( \log(2 s \mu(\cA)) \cdot \log^2(s \log(2N/s)) \log(N) + \log(\epsilon^{-1}) \right ).
}
In other words, the measurement condition scales linearly in $s$, up to the coherence $\mu(\cA)$ and a polylogarithmic factor in $s$, $\mu(\cA)$, $N$ and $\epsilon^{-1}$.

\rem{
[Assumption (iii) of Corollary \ref{cor:main-res-deltaU-subspaces-II} always holds]
\label{rem:ass-iii-holds}
Consider the setup of \S \ref{s:setup}, where $\bbU \subseteq \bbX_0$ satisfies (i) and (ii) of Corollary \ref{cor:main-res-deltaU-subspaces-I}. Then
\bes{
\{ u \in \Delta \bbU : \nm{u}_{\bbX} \leq 1\} \subseteq \bigcup^{d}_{i=1} \{ v \in \bbV_i : \nm{v}_{\bbX} \leq 1 \}.
}
Let $\{ v^{(i)}_{j} \}^{n_i}_{j=1}$ be an orthonormal basis of $\bbV_i$, where $n_i = \dim(\bbV_i) \leq n$ by assumption, and define $\nm{c}^{*}_{1} = \sum^{n_i}_{j=1} (| \Re(c_i) | + |\Im(c_i)|)$ for $c \in \bbC^{n_i}$. Then
\bes{
\nm{c}^*_1 \leq \sqrt{2} \nm{c}_1 \leq \sqrt{2 n_i} \nm{c}_2 = \sqrt{2 n_i} \nm{v}_{\bbX},
}
for arbitrary $v = \sum^{n_{i}}_{j=1} c_j v^{(i)}_j \in \bbV_i$.
If $\bbW_i = \left \{ \pm \sqrt{2 n_i} v^{(i)}_j , \pm \I \sqrt{2 n_i} v^{(i)}_j : j = 1,\ldots,n_i  \right \}$, we deduce that $\{ v \in \bbV_i : \nm{v}_{\bbX} \leq 1 \} \subseteq \mathrm{conv}(\bbW_i)$ and therefore
\bes{
\{ u \in \Delta \bbU : \nm{u}_{\bbX} \leq 1\} \subseteq \bigcup^{d}_{i=1} \mathrm{conv}(\bbW_i) \subseteq \mathrm{conv}(\bbW),\quad \text{where } \bbW = \bigcup^{d}_{i=1} \bbW_i.
}
This shows that assumption (iii) of Corollary \ref{cor:main-res-deltaU-subspaces-II} always holds whenever assumptions (i) and (ii) of Corollary \ref{cor:main-res-deltaU-subspaces-I} hold. Unfortunately, this argument may be too crude in practice to be useful. Since $M = |\bbW | \leq 4 n d$, this leads to a term $\log(2M)$ that behaves like $\log(8 n d) = \log(4 n ) + \log(2 d)$ in the various bounds in Corollary \ref{cor:main-res-deltaU-subspaces-II}. This scales logarithmically in $d$, rather than double-logarithmically, as in the other log term in $d$ in Corollary \ref{cor:main-res-deltaU-subspaces-II}, and may therefore be problematic. In the classical compressed sensing case, for instance, we have $\log(8 n d ) \lesssim s \log(\E N/s)$ due to \ef{d-val-CS}. Thus, using this estimate would lead to a quadratic scaling in $s$ in the overall measurement bound. Fortunately, in this specific case, we can derive a smaller set $\bbW$  (see \ef{conv-hull-sparsity}) which leads to a linear scaling in $s$.
}

\subsection{Extension to structured sparse models}\label{ss:structured-sparse}

Our main results can be applied to many common structured sparse models. The main hurdle is checking that conditions (i)--(iii) of Corollary \ref{cor:main-res-deltaU-subspaces-II} hold. For the well-known \textit{weighted sparsity} model \cite{rauhut2016interpolation}, \cite[Proof of Lem.\ 5.3]{rauhut2016interpolation} shows that assumption (iii) holds with $| \mathbb{W} | \leq 4 N$. For the \textit{sparsity in levels} model \cite{adcock2021compressive}, assumption (iii) holds with  $| \mathbb{W} | \leq 4 N$ \cite[proof of Lem.\ 13.29]{adcock2021compressive}.
Another common structured sparsity model is the \textit{cosparse} model. This arises in analysis compressed sensing problems \cite{genzel2021analysis,nam2013cosparse,kabanava2015cosparsity}, where the analysis operator is not an orthonormal basis. This model satisfies assumptions (i) and (ii). Moreover, if the analysis operator is a Parseval frame it also satisfies (iii) with $|\bbW| \leq 4 N$. See \cite[proof of Cor.\ 4.2]{krahmer2015compressive}.

For illustration, we now consider the \textit{group sparsity} model \cite{davenport2012introduction,duarte2011structured}. Let $\cG = \{ G_i \}^{P}_{i=1}$ be a partition of $[N]$ into nonoverlapping groups of size $|G_i| = g : = N/P$, which we assume to be an integer, for convenience.
Let $x \in \bbC^N$ and $x_{G_i}$ be the restriction of $x$ to the indices in the $i$th group $G_{i}$. We say that $x$ is \textit{$s$-group sparse} for some $1 \leq s \leq P$ if $| \{ i : x_{G_i} \neq 0 \} | \leq s$.
Now define the set
\bes{
\bbU = \bbU_{\cG,s} = \{ x \in \bbC^N : \text{$x$ is $s$-group sparse} \}.
}
As before, we have $\Delta \bbU = \bbU_{\cG , 2s}$. Assumption (i) therefore holds. Assumption (ii) also holds, since we may write $\Delta \bbU$ as a union of ${P \choose 2s}$ subspaces of dimension $2 s g$, namely,
\bes{
\Delta \bbU = \bigcup_{\substack{S \subseteq [P] \\ |S| = 2 s}} \{ x \in \bbC^N : \mathrm{supp}_{\cG}(x) \subseteq S \},\qquad \text{where }\mathrm{supp}_{\cG}(x) = \{ i \in [P] : x_{G_i} \neq 0 \}. 
}
We now verify assumption (iii). By a short argument, 
\bes{
\nm{x}^*_1  : = \sum^{N}_{i=1} ( |\Re(x_i) | + | \Im(x_i) | ) 
 \leq \sqrt{2} \sum_{i \in S} \nm{x_{G_i}}_1 \sqrt{4 s g} \nm{x}_2,\quad \forall x \in \Delta \bbU.
}
Hence, assumption (iii) holds with $|\bbW| = 4 N$ and $\bbW =   \{ \pm \sqrt{4 s g} e_i , \pm  \I \sqrt{4 s g} e_i : i = 1,\ldots,N \}$. 
Now consider the isotropic vector model of Example \ref{ex:comp-sens-isotropic}. Short arguments give that
\bes{
\Phi(S(\Delta \bbU) \cup \bbW ; \bar{\cA}) \leq 4\mu(\cA) s g
}
in this case, where $\mu(\cA)$ is as in \ef{cs-coherence}. Corollary \ref{cor:main-res-deltaU-subspaces-II} now gives the measurement condition
\bes{
m \gtrsim \mu(\cA) \cdot s \cdot g \cdot \left [ \log(2 s g \mu(\cA) ) \cdot \log^2[s g \log(2P/s)]\cdot \log(N)  + \log(1/\epsilon) \right ].
}
This generalizes \ef{goodr-sunglasses}, which corresponds to the case $g = 1$. Notice that the number of unknowns of an $s$-group sparse signal is $s g$. Hence the scaling $s g$ is as expected.

\section{Application to compressed sensing with generative models}\label{s:applic-gen-CS}

In this section, we demonstrate how our main results can be applied to the model classes of Example \ref{ex:generative-models}. In particular, we now establish the first results for Lipschitz maps with general measurements.

\subsection{Recovery guarantee for generative models}

We consider the discrete setting, where $\bbX_0 = \bbX = \bbR^N$ and $\bbZ = \bbR^k$.
Our main result in this section is the following.

\thm{
[Recovery guarantee for generative models]
\label{thm:lipschitz-model-classes}
Consider the setup of \S \ref{s:setup}, where $\bbX_0 = \bbX = \bbR^N$ with the Euclidean inner product and norm. Let $F : \bbR^{k} \rightarrow \bbR^N$ be
\begin{enumerate}[(i)]
\item Lipschitz with constant at most $L > 0 $, i.e., $\nm{F(z) - F(z')}_{\ell^2} \leq L \nm{z - z'}_{\ell^2}$ for all $ z , z' \in \bbR^k$,
\item positively homogeneous, i.e., $F(t z) = t F(z)$ for all $t \geq 0$ and $z \in \bbR^k$,
\end{enumerate}
Let $x \in \bbR^N$ and suppose that there exists $\eta \geq 0$, $0 \leq \chi \leq 1/9$ and $\iota_x \geq 0$ such that
 \begin{enumerate}[(a)]
\item $\nm{x - x^*}_{\ell^2} \leq \chi \nm{x}_{\ell^2}$ for some $x^* = F(z^*)$ with $\nm{z^*}_{\ell^2} \leq \eta$,
\item $(1-\chi) \nm{x}_{\ell^2} \leq \iota_x \leq (1+\chi) \nm{x}_{\ell^2}$.
\end{enumerate}
Let $\bbU_0 = \mathrm{Ran}(F) = \{ u = F(z) : z \in \bbR^k \}$ and
\be{
\label{U-Lipschitz-map-def}
\bbU = \left \{ u = F(z) : \nm{z}_{\ell^2} \leq 2\eta , \nm{u}_{\ell^2} \geq \left (\frac{1+\chi}{1-\chi} \right )^2 \iota_x  \right \} \subseteq \bbU_0,
}
and suppose that either
\be{
\label{meas-cond-lipschitz-alt-1}
m \gtrsim  \alpha^{-1} \cdot \Phi(S(\Delta \bbU_0) ; \bar{\cA}) \cdot \left [ k \log \left ( 2 \left ( 1 + \frac{L \eta \Phi(S(\Delta^2 \bbU) ; \bar{\cA})}{\sqrt{\alpha} \chi \iota_x \Phi(S(\Delta \bbU) ; \bar{\cA}) }  \right ) \right ) + \log(2/\epsilon) \right ]
}
or
\be{
\label{meas-cond-lipschitz-alt-2}
m \gtrsim  \alpha^{-1} \cdot \Phi(S(\Delta^2 \bbU_0) ; \bar{\cA}) \cdot \left [ k \log \left ( 2 \left ( 1 + \frac{L \eta}{\sqrt{\alpha} \chi \iota_x  }  \right ) \right ) + \log(2/\epsilon) \right ].
}
Let $\theta \geq \nm{x}_{\bbX}$ and $\gamma \geq 0$. Then
\be{
\label{err-bd-lipschitz-alt}
\bbE \nm{x - \check{x}}^2_{\ell^2} \lesssim \frac{\beta}{\alpha} \chi^2 \nm{x}^2_{\ell^2} + \theta^2 \epsilon+ \frac{\gamma^2}{\alpha} + \frac{\nm{e}^2_{\overline{\bbY}}}{\alpha},
}
for any $(\gamma,\theta)$-minimizer $\check{x}$ of \ef{least-squares-problem} with noisy measurements \ef{noisy-meas}. 
}
The key facet of the measurement conditions \ef{meas-cond-lipschitz-alt-1}--\ef{meas-cond-lipschitz-alt-2} is that they scale linearly in the latent space dimension $k$, which is the intrinsic complexity of the model class in this case. Hence Theorem \ref{thm:lipschitz-model-classes} shows that recovery using generative models from a near-optimal number of measurements is possible, whenever the variation is small. We discuss this latter issue in \S \ref{ss:gen-mod-rec-unitary} for different types of measurements, as well as describing a theoretically-optimal active learning strategy that follows directly from Theorem \ref{thm:lipschitz-model-classes}.  Note \ef{meas-cond-lipschitz-alt-1} and \ef{meas-cond-lipschitz-alt-2} involve $\Delta \bbU_0$ and $\Delta^2 \bbU_0$. These are particular advantageous since, as we discuss in Remark \ref{rem:compute-coherences}, we can often compute the variation with respect to these sets.

This theorem is a consequence of Theorem \ref{t:main-res}, and motivates why the latter gives a local guarantee.
As discussed in Example \ref{ex:generative-models}, it is customary to consider the model class $\bbU_0= \mathrm{Ran}(F)$. In order to handle the general types of measurements considered in this work, this theorem considers the restricted class \ef{U-Lipschitz-map-def}. By using this restricted class, we are able to bound the entropy integral appearing in Theorem \ref{t:main-res}.

The class \ef{U-Lipschitz-map-def} is specified by constants $\eta$, $\chi$ and $\iota_{x}$ that satisfy (a) and (b). The first is an upper bound on the allowed latent codes $z$, the second is an accuracy parameter and the third is an approximation to $\nm{x}_{\ell^2}$. Condition (a) says that the unknown $x$ should be approximated up to accuracy $\chi$ with some $x^* \in \mathrm{Ran}(F)$ having latent code at most $\eta$ in norm. Condition (b) states that $\iota_x$ should approximate $\nm{x}_{\ell^2}$ to within accuracy $\chi$. These three constants are then used to define $\bbU$, which consists of elements of $\bbU_0 = \mathrm{Ran}(F)$ whose norms exceed $\iota_x$ by $\ord{\chi}$ for small $\chi$ and whose latent codes are at most $2 \eta$ in norm. 

The influence of constant $\eta$ in Theorem \ref{thm:lipschitz-model-classes} is mild, as it only appears logarithmically in the measurement conditions \ef{meas-cond-lipschitz-alt-1} and \ef{meas-cond-lipschitz-alt-2}. Consequently, there is little penalty for choosing a larger $\eta$. Moreover, $\eta$ typically does not need to be taken too large. Recall that a generative neural network $F$ has an associated probability measure $\rho$ on the latent space $\bbR^k$, which is typically chosen as $\rho = \cN(0,I)$. Latent vectors that yield good approximations are not expected to come from the tail of this distribution. Using standard results on Gaussian concentration, we observe that
\bes{
\eta \geq \sqrt{k} + \sqrt{2 \log(1/\epsilon)} \quad \Rightarrow \quad \rho \left ( \left \{ z : \nm{z}_{\ell^2} \leq \eta \right \} \right ) \geq 1-\epsilon.
}
Thus, choosing $\eta \approx \sqrt{k}$ is a reasonable choice in practice.

The accuracy constant $\chi$ could, in practice, be chosen as some desired tolerance. Larger $\chi$ translates into a worse error bound \ef{err-bd-lipschitz-alt}, while if $\chi$ is too small than (a) may not hold, meaning that recovery is no longer guaranteed. Finally, the constant $\iota_x$ should estimate $\nm{x}_{\ell^2}$. It is often possible to do this using the measurements themselves. Indeed, if the family $\bar{\cA}$ is jointly isotropic, i.e., \ef{nondegeneracy} holds with $\alpha = \beta$, as is often the case in practice, then the $\ell^2$-norm $\nm{b}_{\ell^2}$ of the measurements should provide a good approximation to $\nm{x}_{\ell^2}$ under a less stringent measurement condition than that required to recover $x$ itself. 

Finally, we note that Theorem \ref{thm:lipschitz-model-classes} also requires (ii). This holds for ReLU generative neural networks (with no biases, as is customary \cite{berk2022coherence}) as well as certain other types of generative models. It can be relaxed to the condition that $F(t z) = \phi(t) F(z)$, $\forall t \geq 0,\ z \in \bbR^k$,
and some increasing function $\phi : [0,\infty) \rightarrow [0,\infty)$ with $\phi(0) = 0$. For succinctness we do not consider this extension.

\subsection{Recovery guarantees for randomly-subsampled unitary matrices}\label{ss:gen-mod-rec-unitary}

Theorem \ref{thm:lipschitz-model-classes} applies to the general measurement model introduced in \S \ref{s:setup}.
As mentioned, \cite{berk2022coherence,berk2023model} consider measurements drawn from the rows a unitary matrix $U$, which, as described in Examples \ref{ex:comp-sens-sub-unitary} and \ref{ex:bernoulli-model}, is a special case of our general model. We now consider this problem. Let $U \in \bbC^{N \times N}$ be unitary and $u_i = U^* e_i$ for $i = 1,\ldots,N$. Following \cite[Defn.\ 1.4]{berk2023model}, we say that the \textit{local coherences} of $U$ with respect to a set $\bbV \subseteq \bbC^N$ are the entries of the vector $\sigma = (\sigma_i)^{N}_{i=1}$, where
\be{
\label{sigma-i-def}
\sigma_i = \sup_{\substack{v \in \bbV \\ \nm{v}_{\ell^2} = 1}} | u^*_i v |^2,\quad i = 1,\ldots,N.
}
(Note that \cite[Defn.\ 3]{berk2022coherence} uses the notation $\alpha$. We use $\sigma$ as $\alpha$ is already used in the definition of nondegeneracy. We also change \cite[Defn.\ 3]{berk2022coherence} by squaring, as it simplifies the various expressions that follow.) We now show how the local coherence relates to a special case of the variation (Definition \ref{def:variation}). Let $\bar{\cA}$ be the isotropic family of the subsampled unitary matrix model of Example \ref{ex:comp-sens-sub-unitary}. Then $\Phi(S(\bbV) ; \bar{\cA})$ is readily seen to be
\be{
\label{variation-sigmas}
\Phi(S(\bbV) ; \bar{\cA}) = \max_{i=1,\ldots,N} \sup \left\{ \frac{| u^*_i v|^2}{\pi_i \nm{v}^2_{\ell^2}} : v \in \bbV,\ v \neq 0 \right \} = \max_{i=1,\ldots,N} \left \{ \frac{\sigma_i}{\pi_i} \right \}.
}
Using this and recalling that $\alpha = 1$ in this example, we immediately deduce the following.

\cor{
[Generative models and subsampled unitary matrices]
\label{cor:gen-mod-sub-unitary}
Consider $F : \bbR^k \rightarrow \bbR^N$ as in Theorem \ref{thm:lipschitz-model-classes}, $\bbU_0 = \mathrm{Ran}(F)$ and consider the randomly-subsampled unitary matrix model of Example \ref{ex:comp-sens-sub-unitary}. Let $\sigma = (\sigma_i)^{N}_{i=1}$ be the local coherences of $\Delta \bbU_0$ with respect to $\bbU - \bbU$ and $\tilde{\sigma} = (\tilde{\sigma}_i)^{N}_{i=1}$ be the local coherences of $U$ with respect to $\Delta^2 \bbU_0$. Then the measurement conditions \ef{meas-cond-lipschitz-alt-1} and \ef{meas-cond-lipschitz-alt-2} are implied by
\bes{
m \gtrsim C \cdot \left [ k \cdot \log \left ( 2 \left ( 1 + \frac{L \eta \widetilde{C}}{ \chi \iota_x C} \right ) \right ) + \log(2/\epsilon) \right ]
} 
and
\bes{
m \gtrsim \widetilde{C} \cdot \left [ k \cdot \log \left ( 2 \left ( 1 + \frac{L \eta}{ \chi \iota_x } \right ) \right ) + \log(2/\epsilon) \right ],
}
respectively, where $C =  \max_{i=1,\ldots,N} \left ( \frac{\sigma_i}{\pi_i} \right )$ and $\widetilde{C} = \max_{i=1,\ldots,N} \left ( \frac{\tilde{\sigma}_i}{\pi_i} \right ) $.
}

Using this result, we next derive optimized sampling strategies based on the local coherences. We do this by choosing the probabilities $\pi_i$ to minimize either the constant $C$ (i.e., $\pi_i \propto \sigma_i$) or the constant $\widetilde{C}$ (i.e., $\pi_i \propto \tilde{\sigma}_i$). This immediately yields the following result.

\cor{
[Optimal sampling for generative models]
\label{cor:gen-mod-unitary-sub-opt}
Consider the setup of Corollary \ref{cor:gen-mod-sub-unitary}. If $\pi_i = \frac{\sigma_i}{\nm{\sigma}}_{\ell^1}$, $i = 1,\ldots,N$, then the measurement condition \ef{meas-cond-lipschitz-alt-1} is implied by
\be{
\label{gen-mod-unitary-cond-opt-1}
m \gtrsim \nm{\sigma}_{\ell^1} \cdot \left [ k \cdot \log \left ( 2 \left ( 1 + \frac{L \eta }{ \chi \iota_x } \max_{i=1,\ldots,N} \left \{ \frac{\tilde{\sigma}_i^2 }{ \sigma_i } \right \}\right ) \right ) + \log(2/\epsilon) \right ]
}
and if  $\pi_i = \frac{\tilde\sigma_i}{\nm{\tilde\sigma}_{\ell^1}}$, $i = 1,\ldots,N$, then the measurement condition \ef{meas-cond-lipschitz-alt-2} is implied by
\be{
\label{gen-mod-unitary-cond-opt-2}
m \gtrsim \nm{\tilde{\sigma}}_{\ell^1} \cdot \left [ k \cdot \log \left ( 2 \left ( 1 + \frac{L \eta }{\chi \iota_x } \right ) \right ) + \log(2/\epsilon) \right ]. 
}
}

This result describes a theoretically-optimal active learning strategy for compressed sensing with generative models in the case of random sampling with an arbitrary unitary matrix $U$. This result and Corollary \ref{cor:gen-mod-sub-unitary} are similar to those in \cite{berk2023model}. However, \cite{berk2023model} considers only ReLU generative NNs, and the logarithmic factors depend on the widths and depth of the NN. Conversely, the above results apply to generative models defined by arbitrary Lipschitz maps and depend on the Lipschitz constant $L$. As noted, Corollary \ref{cor:gen-mod-unitary-sub-opt} provides an active learning strategy.  It differs from the standard setting of Examples \ref{ex:function-regression} and \ref{ex:christoffel-function} in that the underlying object is a vector, not a function, and the measurements are not pointwise samples. It does, however, fall into the general active learning framework of \cite{adcock2023cs4ml}. Notice that the local coherences $\sigma$ and $\tilde\sigma$ are a specific instance (corresponding to this problem) of what was termed the \textit{generalized Christoffel function} in \cite{adcock2023cs4ml}. Hence Corollary \ref{cor:gen-mod-unitary-sub-opt} is a type of Christoffel sampling of the form considered in \cite{adcock2023cs4ml}, albeit with non-pointwise samples and a highly nonlinear model class.

\rem{
[Computing local coherences]
\label{rem:compute-coherences}
To implement the sampling strategies introduced in Corollary \ref{cor:gen-mod-unitary-sub-opt} we need to compute the local coherences $\sigma_i$ or $\tilde{\sigma}_i$. Consider the former (the latter can be addressed in the same way). Then
\bes{
\sigma_i = \max \left \{ \frac{| u^*_i (F(z_1) - F(z_2) ) |^2}{\nm{F(z_1) - F(z_2) }^2_{\ell^2}} : z_1,z_2 \in \bbR^k,\ F(z_1) \neq F(z_2) \right \}.
}
One option is to numerically solve this maximization problem via (stochastic, accelerated) gradient ascent. However, as shown in \cite{adcock2023cs4ml,berk2023model}, an acceptable approximation can often be obtained in a simpler (and faster) way, by setting
\be{
\label{broken-panda}
\sigma_i \approx \sigma'_i : = \max_{j = 1,\ldots,K} \frac{| u^*_i (F(z_{1,j}) - F(z_{2,j}) ) |^2}{\nm{F(z_{1,j}) - F(z_{2,j}) }^2_{\ell^2}},
}
where $K \gg 1$, $z_{1,1},\ldots,z_{1,K}, z_{2,1},\ldots,z_{2,K} \sim_{\mathrm{i.i.d.}} \rho$ and $\rho$ is the probability measure on the latent space $\bbR^k$ associated with the generative model $F$.
}
The previous results use sampling with replacement. We next show that the same result can be obtained without replacement by using Bernoulli sampling (Example \ref{ex:bernoulli-model}).

\cor{[Optimal Bernoulli sampling for generative models]
\label{cor:gen-mod-unitary-sub-bernoulli}
Let $F : \bbR^k \rightarrow \bbR^N$ be as in Theorem \ref{thm:lipschitz-model-classes}, $\bbU_0 = \mathrm{Ran}(F)$ and consider the Bernoulli sampling model of Example \ref{ex:bernoulli-model}. Let $\sigma_i$ and $\tilde\sigma_i$ be as in Corollary \ref{cor:gen-mod-unitary-sub-opt}. Then the following hold.
\begin{enumerate}[(i)]
\item Let $\pi_i = \min \left \{ 1/m , \sigma_i / c_{\pi} \right \}$, $i = 1,\ldots,N$,
where $c_{\pi}$ is the unique solution in $[0,\infty)$ to the nonlinear equation $\sum^{N}_{i=1} \min \left \{ 1/m , \sigma_i / c \right \} = 1$.
Then \ef{meas-cond-lipschitz-alt-1} is implied by the condition
\be{
\label{gen-mod-unitary-cond-opt-1-bernoulli}
m \gtrsim \nm{\sigma}^2_{\ell^2} \cdot \left [ k \cdot \log \left ( 2 \left ( 1 + \frac{L \eta }{ \chi \iota_x } \max_{i=1,\ldots,N} \left \{ \frac{\tilde{\sigma}_i^2 }{ \sigma_i } \right \}\right ) \right ) + \log(2/\epsilon) \right ] .
}
\item Let $\pi_i = \min \left \{ 1/m , \tilde\sigma_i / c_{\pi} \right \}$, $i = 1,\ldots,N$,
where $c_{\pi}$ is the unique solution in $[0,\infty)$ to the nonlinear equation $\sum^{N}_{i=1} \min \left \{ 1/m , \tilde\sigma_i / c \right \} = 1$.
Then \ef{meas-cond-lipschitz-alt-2} is implied by the condition
\be{
\label{gen-mod-unitary-cond-opt-2-bernoulli}
m \gtrsim \nm{\tilde{\sigma}}^2_{\ell^2} \cdot \left [ k \cdot \log \left ( 2 \left ( 1 + \frac{L \eta }{\chi \iota_x  } \right ) \right ) + \log(2/\epsilon) \right ].
}
\end{enumerate}
}

Note that \ef{gen-mod-unitary-cond-opt-1-bernoulli} and \ef{gen-mod-unitary-cond-opt-2-bernoulli} are identical to \ef{gen-mod-unitary-cond-opt-1} and \ef{gen-mod-unitary-cond-opt-2}, respectively. Hence this result shows that the sampling without replacement can be achieved under the same sample complexity bounds as sampling with replacement.

\rem{
\label{rem:why-nonconstant}
The proof of this corollary also demonstrates why it is important to define the variation as a maximum over the nonconstant distributions only (recall the discussion below Definition \ref{def:variation-collection}). The resulting measurement conditions could not be obtained had the variation been defined over all distributions.
}

\subsection{Extension to block sampling}\label{ss:block-sampling}

A desirable aspect of the general framework of \S \ref{s:setup} is that we can easily derive similar results for compressed sensing with generative models under a more general `block' sampling model \cite{bigot2016analysis,boyer2019compressed}. Here, each measurement corresponds to a block of rows of a unitary matrix $U$, rather than a single row. This model is highly relevant in applications such as MRI, where it is generally impossible to measure individual frequencies (i.e., individual rows). See \cite{adcock2021compressive,mcrobbie2006mri}.

We now formalize this setup. Note that this was previously done in \cite[\S C.1]{adcock2023cs4ml} for the special case where $U$ was the 3D Discrete Fourier Transform (DFT) matrix and each block corresponded to a horizontal line of $k$-space values. Here we consider the general setting.
Let $P_1,\ldots,P_M$ be a (potentially overlapping) partition of $[N]$, $p_i = | P_i |$, $p = \max_{i=1,\ldots,M} \{ p_i\}$ and
\bes{
r_j = | \{ i : j \in P_i \} |,\quad j = 1,\ldots,N,
}
be the number of partition elements to which the integer $j$ belongs. Let $\pi = (\pi_1,\ldots,\pi_M)$ be a discrete probability distribution on $[M]$ and define $B_i \in \cB(\bbC^N , \bbC^p)$ by
\bes{
B_i x = \frac{1}{\sqrt{\pi_i}} \begin{bmatrix} ((U x)_j / \sqrt{r_j})_{j \in P_i} \\ 0_{p -p_i} \end{bmatrix},\quad x \in \bbC^N,
}
where $0_{p-p_i}$ is the vector of zeros of size $(p-p_i) \times 1$. Now define the distribution $\cA$ of bounded linear operators in $\cB(\bbC^N , \bbC^p)$ by $A \sim \cA$ if
\bes{
\bbP(A = B_i) = \pi_i,\quad i = 1,\ldots,M.
}
We readily deduce that $\cA$ is isotropic. Indeed,
\bes{
\bbE_{A \sim \cA} \nm{A x}^2_{\ell^2} = \sum^{M}_{i=1} \pi_i \frac{1}{\pi_i} \sum_{j \in P_i} \frac{|(U x)_j |^2}{r_j} = \nm{U x}^2_{\ell^2} = \nm{x}^2_{\ell^2},\quad \forall x \in \bbC^N.
}
Letting $\cA_i = \cA$, $\forall i$, gives the desired block sampling model. Note that each measurement $b_i \in \bbC^{p}$ in \ef{noisy-meas} is associated to a block of rows of $U$ (up to scaling factors to account for overlaps) from one of the partition elements, which is chosen randomly according to $\pi$.

The setting considered corresponds to the case $N = M$ and $P_i = \{i \}$ for $i = 1,\ldots,N$. Much as in that case, we can readily derive a measurement condition for block sampling. The only change we need to make is to redefine the local coherences $\sigma_i$ in \ef{sigma-i-def} by the following, which we term the \textit{block} local coherences:
\be{
\label{sigma-i-def-block}
\sigma_i = \sup_{\substack{v \in \bbV \\ \nm{v}_{\ell^2} = 1}}  \sum_{j \in P_i} \frac{|u^*_j v |^2}{r_j},\quad i = 1,\ldots,M.
}

\cor{
[Generative models and block sampling]
\label{cor:gen-mod-sub-block}
Corollaries \ref{cor:gen-mod-sub-unitary}--\ref{cor:gen-mod-unitary-sub-bernoulli} also hold for the block sampling model with $N$ replaced by $M$ and taking \ef{sigma-i-def-block} as the definition of the $\sigma_i$ rather than \ef{sigma-i-def}.
}

This block model was used recently in \cite{adcock2023cs4ml} in the context of MRI reconstruction with generative models (see \cite[App.\ C]{adcock2023cs4ml}). Note that the local coherences \ef{sigma-i-def-block} are identical to the generalized Christoffel functions defined therein. Here we generalize this setup and provide theoretical guarantees. On the practical side, we note that the block local coherences case \ef{sigma-i-def-block} can be estimated empirically, much as in \ef{broken-panda}.

\section{Technical ingredients for the proofs}\label{ss:cov-num-prop}

The rest of the paper is devoted to the proofs of the main results. For these, we require several standard properties of covering numbers (Definition \ref{def:cov-num}). See, e.g., \cite[Lem.\ 13.22, 13.23]{adcock2021compressive}.
\begin{enumerate}[(A)]
\item $\cN(J \cup K , d , t) \leq \cN(J,d,t) + \cN(K,d,t)$.
\item $\cN(J,d,t) \leq \cN(K,d,t/2)$ whenever $J \subseteq K$.
\item $\cN(K,ad,t) = \cN(K,d,t/a)$ for all $a > 0$.
\item If $d'$ is another pseudometric on $\bbN$ with $d'(x,y) \leq d(x,y)$, $\forall x,y \in \bbM$, then $\cN(K,d',t) \leq \cN(K,d,t)$.
\item If $d$ is induced by a seminorm, then $\cN(a K,d,t) = \cN(K,d,t/a)$ for all $a > 0$.
\item If $d = \nms{\cdot}$ is a norm on $\bbM = \bbR^n$ and $K$ is a subset of the unit ball $\{ x \in \bbR^n : \nm{x} \leq 1\}$, then $\cN(K,\nms{\cdot},t) \leq (1+2/t)^n$.
\end{enumerate}
We also require the following (see \cite[Lem.\ C.9]{foucart2013mathematical}).

\lem{
\label{lem:integral-bound}
For any $\alpha > 0$, $
\int^{\alpha}_{0} \sqrt{\log(1+t^{-1}) } \D t \leq \alpha \sqrt{\log(\E(1+\alpha^{-1}))}$.
}
It is convenient for our purposes to reformulate this result by a simple change of variables:
\be{
\label{log-integral-bd}
\int^{1/2}_{0} \sqrt{\log(1+\beta t^{-1}) } \D t \leq \frac12 \sqrt{\log(\E(1+2 \beta))},\quad \forall \beta > 0.
}

\section{Proof of Theorem \ref{t:main-res}}\label{s:proof-main}

Consider a realization of the $A_i$. We say that \textit{empirical nondegeneracy} holds over a set $\bbU \subseteq \bbX_0$ with constants $0 < \alpha' \leq \beta' < \infty$ if
\be{
\label{empirical-nondegeneracy}
\alpha' \nm{v}^2_{\bbX} \leq \frac{1}{m} \sum^{m}_{i=1} \nm{A_i(v)}^2_{\bbY_i} \leq \beta' \nm{v}^2_{\bbX},\quad \forall v \in \bbV.
}
Recall from the notation introduced in \S \ref{s:setup} this can be equivalently written as
\bes{
\alpha' \nm{v}^2_{\bbX} \leq \frac1m \nm{\bar{A}(v)}^2_{\overline{\bbY}} \leq \beta' \nm{v}^2_{\bbX},\quad \forall v \in \bbV.
}
Note that in the classical compressed sensing setup (see Example \ref{ex:comp-sens-isotropic}), this equivalent to the measurement matrix $\frac{1}{\sqrt{m}} A$ satisfying the Restricted Isometry Property (RIP). 

\subsection{Empirical nondegeneracy suffices for recovery}

The following lemma shows that empirical nondegeneracy implies recovery.

\lem{
[Empirical nondegeneracy implies an error bound]
\label{lem:main-err-bound}
Let $x^* \in \bbX_0$ and $A_i \in \cB(\bbX_0,\bbY_i)$, $i = 1,\ldots,m$, be such that \ef{empirical-nondegeneracy} holds over the set $\bbU - \{ x^* \} : = \{ u - x^* : u \in \bbU \}$ with constants $0 < \alpha' \leq \beta' < \infty$. Let $x \in \bbX_0$, $\gamma \geq 0$ and $\hat{x} \in \bbU$ be a $\gamma$-minimizer of \ef{least-squares-problem} based on noisy measurements \ef{noisy-meas}. Then, for all $u \in \bbU$,
\be{
\label{main-err-bound-u*}
\nm{x - \hat{x}}_{\bbX} \leq \frac{1}{\sqrt{\alpha'}} \nms{\frac{1}{\sqrt{m}} \bar{A}(x - x^*)}_{\overline{\bbY}} + \frac{1}{\sqrt{\alpha'}} \nms{\frac{1}{\sqrt{m}} \bar{A}(x - u)}_{\overline{\bbY}} + \nm{x -x^* }_{\bbX}  + \frac{2}{\sqrt{\alpha'}} \nm{\bar{e}}_{\overline{\bbY}} + \frac{\gamma}{\sqrt{\alpha'}}.
}
}
\prf{
By the triangle inequality, \ef{empirical-nondegeneracy} and the fact that $\hat{x}$ is a $\gamma$-minimizer, we have
\eas{
\nm{\hat{x} - x}_{\bbX} & \leq \nm{\hat{x} - x^*}_{\bbX} + \nm{x - x^*}_{\bbX}
\\
& \leq 1/\sqrt{\alpha'} \nms{\frac{1}{\sqrt{m}} \bar{A}(\hat{x}-x^*)}_{\overline{\bbY}} + \nm{x-x^*}_{\bbX} 
\\
& \leq 1/\sqrt{\alpha'} \nms{\frac{1}{\sqrt{m}} \bar{A}(\hat{x}) - \bar{b}}_{\overline{\bbY}} + 1/\sqrt{\alpha'}  \nms{\frac{1}{\sqrt{m}}  \bar{A}(x^*) - \bar{b} }_{\overline{\bbY}} + \nm{x - x^*}_{\bbX}
\\
& \leq \gamma / \sqrt{\alpha'} + 1/\sqrt{\alpha'} \nms{\frac{1}{\sqrt{m}}\bar{A}(u) - \bar{b}}_{\overline{\bbY}} + 1/\sqrt{\alpha'}  \nms{\frac{1}{\sqrt{m}}\bar{A}(x^*) - \bar{b}}_{\overline{\bbY}} + \nm{x - x^*}_{\bbX}
\\
& \leq \gamma / \sqrt{\alpha'} + 1/\sqrt{\alpha'}  \nms{\frac{1}{\sqrt{m}} \bar{A}(x-u)}_{\overline{\bbY}} + 1/\sqrt{\alpha'}  \nms{\frac{1}{\sqrt{m}}\bar{A}(x-x^*)}_{\overline{\bbY}}
\\
&~~~+2/\sqrt{\alpha'} \nm{\bar{e}}_{\overline{\bbY}} + \nm{x - x^*}_{\bbX}
}
for any $u \in \bbU$.
This gives the result.
}

Notice that this result only requires the lower inequality in \ef{empirical-nondegeneracy}. The upper inequality will be used later in the derivation of the error bound in expectation.
Consequently, the remainder of the proofs are devoted to deriving measurement conditions under which \ef{empirical-nondegeneracy} holds, then using this and the above lemma to derive error bounds in expectation.

\subsection{Measurement conditions for empirical nondegeneracy}

\thm{
\label{thm:main-res-new}
Consider the setup of \S\ref{s:setup}. Let $0 < \delta,\epsilon < 1$ and $\bbU$ be a subset of a finite-dimensional subspace of $\bbX_0$.
Suppose that, for some $0 < \epsilon < 1$ and $q \geq 1$,
\bes{
m \gtrsim \delta^{-2} \cdot \upsilon^{2+\frac2q} \cdot \left [  \left( \esssup_{\bar{A} \sim \bar{\cA}} \int^{1/2}_{0} \sqrt{\log(2 \cN(\widetilde{S}(\bbU) , \nms{\cdot}_{q,\bar{A}} , \upsilon t  ) ) } \D t  \right)^2  + \log(2/\epsilon) \right ],
}
where $\upsilon = \sqrt{\Phi(S(\bbU) ; \bar{\cA})  / \alpha}$.
Then, with probability at least $1-\epsilon$, \ef{empirical-nondegeneracy} holds for $\bbU$ with constants $\alpha' = (1-\delta) \alpha$ and $\beta' = (1+\delta) \beta$. 
}

To prove this result, it suffices to show that the random variable
\be{
\label{delta_U_def}
\delta_{\bbU} = \sup_{u \in \widetilde{S}(\bbU)} \left | \frac{1}{m} \sum_{i \in \cI} \nm{A_i(u)}^2_{\bbY_i}- 1 \right |
}
satisfies $\delta_{\bbU} \leq \delta$ with probability at least $1-\epsilon$. Following a standard route, we do this by first bounding $\bbE(\delta_{\bbU})$ and then by estimating the probability that $\delta_{\bbU}$ deviates from $\bbE(\delta_{\bbU})$.

We commence with the expectation bound.  Recall that a \textit{Rademacher} random variable is a random variable that takes the values $+1$ and $-1$ with probability $\frac12$. 

\lem{
\label{lem:exp-symmetrize}
Consider the setup of Theorem \ref{thm:main-res-new}.
Let $\cI$ be the index set \ef{index-set-I} and $\{ \epsilon_{i}\}_{i \in \cI}$ be independent Rademacher random variables that are also independent of the random variables $\{ A_i \}^{m}_{i=1}$. Then 
\bes{
\bbE_{\bar{\cA}}(\delta_{\bbU}) \leq 2 m^{-1} \bbE_{\bar{\cA}} \bbE_{\overline{\epsilon}} \sup_{u \in \widetilde{S}(\bbU)} \left | \sum_{i \in \cI} \epsilon_{i} \nm{A_i(u)}^2_{\bbY_i} \right |.
}
Here $\bbE_{\bar{\cA}}$ denotes expectation with respect to the variables $\{ A_i\}^{m}_{i=1}$ and $\bbE_{\overline{\epsilon}}$ denotes expectation with respect to the $\{ \epsilon_i \}_{i \in \cI}$.
}
\prf{
Since
\be{
\label{u-SU-norm}
1 = \tnm{u}^2_{\bbX} = \frac1m \sum^{m}_{i=1} \bbE_{A_i \sim \cA_i} \nm{A_i(u)}^2_{\bbY_i}  ,\quad \forall u \in \widetilde{S}(\bbU),
}
we have
\be{
\label{delta-U-I-def}
\begin{split}
\delta_{\bbU} & = \sup_{u \in \widetilde{S}(\bbU)} \left |  \frac{1}{m} \sum^{m}_{i=1} \left (  \nm{A_i(u)}^2_{\bbY_i}- \bbE_{A_i \sim \cA_i}  \nm{A_i(u)}^2_{\bbY_i}\right ) \right | 
\\
& = \sup_{u \in \widetilde{S}(\bbU)} \left |  \frac{1}{m} \sum_{i \in \cI} \left (  \nm{A_i(u)}^2_{\bbY_i}- \bbE_{A_i \sim \cA_i}  \nm{A_i(u)}^2_{\bbY_i}\right ) \right |\quad \text{a.s.}
\end{split}
}
By assumption, $\bbU \subseteq \overline\bbV$, where $\overline\bbV \subseteq \bbX_0$ is a finite-dimensional vector space. Since $\overline\bbV$ is a vector space, we also have $\widetilde{S}(\bbU) \subseteq \overline\bbV$. Consider $\overline\bbV$ as a Hilbert space with the inner product from $\bbX$. Then the map
\bes{
B_{i} : (\overline\bbV,\nms{\cdot}_{\bbX}) \rightarrow (\bbY_i , \nms{\cdot}_{\bbY_i}),\ v \mapsto A_i(v),
}
is bounded. Indeed,
\bes{
\nm{B_i(v)}_{\bbY_i} \leq \nm{A_i}_{\bbX_0 \rightarrow \bbY_i} \nm{v}_{\bbX_0} \leq c \nm{A_i}_{\bbX_0 \rightarrow \bbY_i} \nm{v}_{\bbX},\quad \forall v \in \overline\bbV.
}
Here, in the first inequality we used the fact that the map $A_i \in \cB(\bbX_0,\bbY_i)$ is bounded by assumption and in the second step, we used the fact that $\overline\bbV$ is finite dimensional and all norms on finite-dimensional vector spaces are equivalent. Therefore, $B_i$ has a unique bounded adjoint $B^*_{i} : (\bbY_i , \nms{\cdot}_{\bbY_i}) \rightarrow (\overline\bbV,\nms{\cdot}_{\bbX})$ and we may write
\bes{
\nm{A_i(v)}^2_{\bbY_i} = \ip{B^*_i B_i(v)}{v}_{\bbX} = \ip{U_i(v)}{v}_{\bbX},\quad \forall v \in \overline\bbV,
}
where $U_{i} = B^*_{i} B_{i}$ is a random variable taking values in $\cB(\overline\bbV)$, the finite-dimensional vector space of bounded linear operators on $(\overline\bbV,\nms{\cdot}_{\bbX})$. Using this we can write
\be{
\label{delta-char-f}
\delta_{\bbU} = m^{-1} \sup_{u \in \widetilde{S}(\bbU)} \left |  \sum_{i \in \cI} \ips{ U_{i}(u) - \bbE \left ( U_{i}(u) \right ) }{u}_{\bbX} \right |= m^{-1} f \left ( \sum_{i \in \cI} ( U_{i} - \bbE (U_{i}) ) \right ),
}
where $f$ is the convex function
\bes{
f : \cB(\overline\bbV) \rightarrow \bbR,\quad U \mapsto \sup_{u \in \widetilde{S}(\bbU)} | \ip{U(u)}{u}_{\bbX} |.
}
We now apply symmetrization to get
\bes{
\bbE_{\bar{\cA}}(\delta_{\bbU}) \leq m^{-1} \bbE_{\bar{\cA}} \bbE_{\overline{\epsilon}} f \left ( 2 \sum_{i \in \cI}\epsilon_{i} U_{i} \right ).
}
Now observe that
\bes{
f \left ( 2 \sum_{i \in \cI} \epsilon_i U_{i} \right ) = 2\sup_{u \in \widetilde{S}(\bbU)} \left | \ips{ \sum_{i \in \cI} \epsilon_{i} U_{i}(u) }{u}_{\bbX} \right |
 = 2 \sup_{u \in \widetilde{S}(\bbU)} \left | \sum_{i \in \cI} \epsilon_{i} \nm{A_i(u)}^2_{\bbY_i} \right |.
}
This completes the proof.
}

The next result involves Dudley's inequality (see, e.g., \cite[Thm.\ 13.25]{adcock2021compressive}). Here we also recall the notation $\widetilde{m} = | \cI |$.

\lem{
\label{lem:exp-dudley-covering-bd}
Consider the setup of Theorem \ref{thm:main-res-new}.
Draw $\bar{A} = (A_1,\ldots,A_m) \sim \bar{\cA}$. Then, with probability one, we have
\be{
\label{exp-covering-dudley}
\bbE_{\overline{\epsilon}} \sup_{u \in \widetilde{S}(\bbU)} \left | \sum_{i \in \cI} \epsilon_{i} \nm{A_i(u)}^2_{\bbY_i} \right | \lesssim \widetilde{m}^{\frac12} R_{p,\bar{A}} \upsilon \int^{1/2}_0 \sqrt{\log(2 \cN(\widetilde{S}(\bbU),\nms{\cdot}_{q,\bar{A}},t)} \D t
}
for all $q \geq 1$, where 
\be{
\label{Rp-dudley-def}
R_{p,\bar{A}} = \sup_{u \in \widetilde{S}(\bbU)} \left (  \frac{1}{\widetilde{m}} \sum_{i \in \cI} \nm{A_i(u)}^{2p}_{\bbY_i}  \right )^{\frac{1}{2p}},
}
and $1 < p \leq \infty$ is such that $1/p + 1/q = 1$, $\upsilon = \sqrt{\Phi(S(\bbU) ; \bar{\cA}) / \alpha}$
and $\nms{\cdot}_{q,\bar{A}}$ is as in \ef{discrete-norm-dudley-def}.
}
\prf{
The left-hand side of \ef{exp-covering-dudley} is the expectaction of the supremum of the absolute value of the Rademacher process $\{ X_u : u \in \widetilde{S}(\bbU) \}$, where
\bes{
X_u =  \sum_{i \in \cI} \epsilon_{i} \nm{A_i(u)}^2_{\bbY_i}.
}
The corresponding pseudometric is
\bes{
d(u,v) = \sqrt{\sum_{i \in \cI} \left (  \nm{A_i(u)}^2_{\bbY_i} -  \nm{A_i(v)}^2_{\bbY_i}\right )^2 },\quad u,v \in \widetilde{S}(\bbU)
}
and
Dudley's inequality implies that
\be{
\label{dudley-applic}
\bbE_{\overline{\epsilon}} \sup_{u \in \widetilde{S}(\bbU)} \left |\sum_{i \in \cI} \epsilon_{i} \nm{A_i(u)}^2_{\bbY_i} \right | \lesssim \int^{\varpi/2}_{0} \sqrt{\log(2 \cN(\widetilde{S}(\bbU),d,z))} \D z,
}
where $\varpi = \sup_{u \in \widetilde{S}(\bbU)} \sqrt{\bbE_{\overline{\epsilon}} | X_u |^2}$. The remainder of the proof involves upper bounding the pseudometric $d$ and the constant $\varpi$.

First, for two sequences $a_i$, $b_i$, observe that
\eas{
\sum_i ( |a_i|^2 - |b_i|^2 )^2 &= \sum_i (|a_i| + |b_i|)^2 (|a_i| - |b_i|)^2 
 \\
 & \leq \left ( \sum_i (|a_i| + |b_i|)^{2p} \right )^{\frac{1}{p}} \left ( \sum_i (|a_i - b_i|)^{2q}  \right )^{\frac1q},
}
where $q$ is such that $1/p+1/q = 1$. Therefore
\bes{
\sqrt{\sum_i ( |a_i|^2 - |b_i|^2 )^2} \leq 2 \max \{ \nm{a}_{2p} , \nm{b}_{2p} \} \nm{a - b}_{2q}.
}
We deduce that, with probability one,
\be{
\label{pseudometric-bound}
d(u,v) \leq 2 \widetilde{m}^{\frac{1}{2}} R_{p,\bar{A}} \nm{u-v}_{q,\bar{A}},\quad \forall u,v \in \widetilde{S}(\bbU).
}
Now consider $\varpi$. Then, defining additionally $X_0 = 0$ (since $0 \notin \widetilde{S}(\bbU)$), we see that 
\bes{
\varpi = \sup_{u \in \widetilde{S}(\bbU)} d(u,0) \leq 2 \widetilde{m}^{\frac{1}{2}} R_{p,\bar{A}} \sup_{u \in \widetilde{S}(\bbU)} \nm{u}_{q,\bar{A}}
}
with probability one.
Now let $u = v / \tnm{v}_{\bbX} \in \widetilde{S}(\bbU)$, where $v \in \bbU$. Then
\bes{
\nm{A_i(v)}^2_{\bbY_i} \leq \Phi(S(\bbU),\cA_i) \nm{v}^2_{\bbX} .
}
Also, recall that $\sqrt{\alpha} \nm{x}_{\bbX} \leq \tnm{x}_{\bbX} \leq \sqrt{\beta} \nm{x}_{\bbX}$, $\forall x \in \bbX_0$. We deduce that
\be{
\label{Ai-over-BtildeU-bd}
\nm{A_i(u)}^2_{\bbY_i} \leq \Phi(S(\bbU); \cA_i) / \alpha,\quad \forall u \in \widetilde{S}(\bbU)
}
with probability one.
This gives
\eas{
\nm{u}_{q,\bar{A}} \leq \upsilon,\quad \forall u \in \widetilde{S}(\bbU)
}
with probability one.
Hence $\varpi \leq 2 \widetilde{m}^{\frac12} R_{p,\bar{A}} \upsilon$ with probability one.

We now combine this with \ef{pseudometric-bound} and properties (C) and (D) of \S \ref{ss:cov-num-prop} to obtain
\bes{
\bbE_{\overline{\epsilon}} \sup_{u \in \widetilde{S}(\bbU)} \left | \frac{1}{m}\sum_{i \in \cI} \epsilon_{i} \nm{A_i(u)}^2_{\bbY_i} \right | \lesssim \int^{\widetilde{m}^{\frac12} R_{p,\bar{A}} \upsilon}_{0}  \sqrt{\log(2 \cN(\widetilde{S}(\bbU) , \nms{\cdot}_{q,\bar{A}} , z / (2 \widetilde{m}^{\frac12} R_{p,\bar{A}}) ))} \D z
}
with probability one.
The result now follows from a change of variables.
}

We are now ready to present the expectation bound.

\thm{[Expectation bound]
\label{thm:exp-bound}
Consider the setup of Theorem \ref{thm:main-res-new}. Suppose that, for some $q \geq 1$,
\bes{
m \gtrsim \delta^{-2} \cdot  \upsilon^{2+\frac{2}{q}}  \cdot   \left( \esssup_{\bar{A} \sim \bar{\cA}} \int^{1/2}_{0} \sqrt{\log(2 \cN(\widetilde{S}(\bbU) , \nms{\cdot}_{q,\bar{A}} , \upsilon t  ) ) } \D t  \right)^2  ,
}
where $\upsilon = \sqrt{\Phi(S(\bbU) ; \bar{\cA})  / \alpha}$.
Then the random variable \ef{delta_U_def} satisfies $\bbE(\delta_{\bbU}) \leq \delta$. 
}

\prf{
The previous two lemmas imply that
\eas{
\bbE_{\bar{\cA}}(\delta_{\bbU}) & \lesssim m^{-1} \widetilde{m}^{\frac12} \upsilon \cdot \bbE_{\bar{\cA}} \left ( R_{p,\bar{A}}  \int^{1/2}_{0} \sqrt{\log(2 \cN(\widetilde{S}(\bbU),\nms{\cdot}_{q,\bar{A}},t)} \D t \right )
\\
& \leq m^{-1} \widetilde{m}^{\frac12} \upsilon \cdot \left ( \esssup_{\bar{A} \sim \bar{\cA}}  \int^{1/2}_{0} \sqrt{\log(2 \cN(\widetilde{S}(\bbU),\nms{\cdot}_{q,\bar{A}},t)} \D t \right ) \bbE_{\bar{\cA}}(R_{p,\bar{A}})
\\
& = : m^{-1} \widetilde{m}^{\frac12} \upsilon \sqrt{C} \cdot \bbE_{\bar{\cA}}(R_{p,\bar{A}}).
}
for all $1 \leq p < \infty$, where $R_{p,\bar{A}}$ is as in \ef{Rp-dudley-def}, 
$1 < q \leq \infty$ is such that $1/p + 1/q = 1$ and the discrete norm $\nms{\cdot}_{q,\bar{A}}$ is as in \ef{discrete-norm-dudley-def}. 
We first consider the term $R_{p,\bar{A}}$. 
Using \ef{Ai-over-BtildeU-bd} and the definition of $R_{p,\bar{A}}$, we see that
\bes{
R_{p,\bar{A}} \leq  \left ( \frac{\Phi(S(\bbU);\bar\cA)}{\alpha} \right )^{\frac12-\frac{1}{2p}}  \left ( \frac{m}{\widetilde{m}} \right )^{\frac{1}{2p}} \sup_{u \in \widetilde{S}(\bbU)}  \left ( \frac{1}{m} \sum_{i \in \cI}  \nm{A_i(u)}^2_{\bbY_i}  \right )^{\frac{1}{2p}}
}
with probability one.
Now consider the sum. Let $u \in \widetilde{S}(\bbU)$ be arbitrary. Then
\eas{
 \frac1m\sum_{i \in \cI}  \nm{A_i(u)}^2_{\bbY_i} &= \frac1m\sum_{i \in \cI} \left( \nm{A_i(u)}^2_{\bbY_i}  - \bbE_{A_i \sim \cA_i} \nm{A_i(u)}^2_{\bbY_i} \right )  +  \frac1m\sum_{i \in \cI}\bbE_{A_i \sim \cA_i} \nm{A_i(u)}^2_{\bbY_i}
 \\
 &   \leq \delta_{\bbU} + 1,
}
where in the second step we used \ef{delta-U-I-def} and \ef{u-SU-norm}. Hence
\bes{
R_{p,\bar{A}}  \leq \left ( \frac{m}{\widetilde{m}} \right )^{\frac{1}{2p}} \left ( \frac{\Phi(S(\bbU);\bar\cA)}{\alpha} \right )^{\frac12-\frac{1}{2p}} (\delta_{\bbU}+1)^{\frac{1}{2p}}
 \equiv \left ( \frac{m}{\widetilde{m}} \right )^{\frac{1}{2p}} \upsilon^{\frac{1}{q}}  (\delta_{\bbU}+1)^{\frac{1}{2p}}
}
with probability one.
Using the fact that $p \geq 1$ and the Cauchy--Schwarz inequality, we deduce that
\be{
\label{E-RpY-in-terms-of-delta}
\bbE_{\bar{\cA}}(R_{p,\bar{A}}) \leq \left ( \frac{m}{\widetilde{m}} \right )^{\frac{1}{2p}} \upsilon^{\frac1q}  \sqrt{\bbE_{\bar{\cA}}(\delta_{\bbU}) + 1} .
}
Combining this with the previous expression, we obtain
\be{
\label{cookies}
\begin{split}
\bbE_{\bar{\cA}}(\delta_{\bbU}) & \lesssim \frac{m^{\frac{1}{2p}-1}}{\widetilde{m}^{\frac{1}{2p}-\frac12} }  \upsilon^{1+\frac1q} \sqrt{C} \sqrt{\bbE_{\bar{\cA}}(\delta_{\bbU}) + 1} 
\leq m^{-\frac12}  \upsilon^{1+\frac1q} \sqrt{C} \sqrt{\bbE_{\bar{\cA}}(\delta_{\bbU}) + 1} ,
\end{split}
}
where in the second step we used the fact that $\widetilde{m} \leq m$. 
Hence, we see that $\bbE_{\bar{\cA}}(\delta_{\bbU}) \leq \delta$, provided
\bes{
m^{-1/2} \upsilon^{1+\frac1q} \sqrt{C} \leq c \delta
}
for suitably small constant $c > 0$. Rearranging now gives the result.
}

Having established an expectation bound, we now bound $\delta_{\bbU}$ in probability.

\thm{
\label{thm:prob-bound}
Consider the setup of Theorem \ref{thm:main-res-new} and let $\delta_{\bbU}$ be as in \ef{delta_U_def}. Suppose that $\bbE(\delta_{\bbU}) \leq \delta / 2$. Then $\delta_{\bbU} \leq \delta$ with probability at least $1-\epsilon$, provided
\bes{
m \gtrsim \delta^{-2} \cdot \alpha^{-1} \cdot \Phi(S(\bbU) ; \bar\cA) \cdot \log(2/\epsilon).
}
}

This theorem is essentially identical to \cite[Thm.\ E.12]{adcock2024unified}. That result assumes $\bbU$ satisfies condition (ii) of Corollary \ref{cor:main-res-deltaU-subspaces-I}. However, this assumption is not required in the proof. For brevity, we omit the proof and now move on to the final arguments for the proof of Theorem \ref{thm:main-res-new}.

\prf{[Proof of Theorem \ref{thm:main-res-new}]
As noted, it suffices to show that $\bbP(\delta_{\bbU} > \delta) \leq \epsilon$, where $\delta_{\bbU}$ is as in \ef{delta_U_def}. Theorem \ref{thm:exp-bound} and the condition on $m$ give that $\bbE(\delta_{\bbU}) \leq \delta/2$. Given this, an application of Theorem \ref{thm:prob-bound} (noting the condition on $m$ once more) then gives the result.
}

\subsection{Proof of Theorem \ref{t:main-res}}

We are now, finally, ready to prove this theorem.

\prf{[Proof of Theorem \ref{t:main-res}]
We follow essentially the same ideas as in \cite[Thm.\ 4.8]{adcock2023cs4ml}.
Let $\delta = 1/2$ (this value is arbitrary), $u^* \in \bbU$ be such that $\nm{x - u^*}_{\bbX} \leq 2 \inf_{u \in \bbU} \nm{x-u}_{\bbX}$ (the constant $2$ is also arbitrary) and $E$ be the event that \ef{empirical-nondegeneracy} holds over $\bbU^*$ with $\alpha' = (1-\delta) \alpha$ and $\beta' = (1+\delta) \beta$. Theorem \ref{thm:main-res-new} (with $\bbU^*$ in place of $\bbU$) and the condition on $m$ implies that $\bbP(E^c) \leq \epsilon$. Now write
\be{
\label{exp-split}
\bbE \nm{x - \check{x}}^2_{\bbX} = \bbE ( \nm{x - \check{x}}^2_{\bbX} | E ) \bbP(E) + \bbE ( \nm{x - \check{x}}^2_{\bbX} | E^c ) \bbP(E^c).
}
Observe that the mapping $\cC : \bbX \rightarrow \bbX, x \mapsto \min \{ 1 , \theta/\nm{x}_{\bbX} \} x$ is a contraction. We also have $\check{x} = \cC(\hat{x})$ and $x = \cC(x)$, where in the latter case, we used the fact that $\theta \geq \nm{x}_{\bbX}$ by assumption.
Consider the first term in \ef{exp-split}. If the event $E$ occurs, then the properties of $\cC$ and Lemma \ref{lem:main-err-bound} give that
\eas{
\nm{x - \check{x}}_{\bbX} & = \nm{\cC(x) - \cC(\hat{x})}_{\bbX} 
\\
& \leq \nm{x - \hat{x}}_{\bbX} 
\\
& \lesssim \frac{1}{\sqrt{\alpha}}  \nms{\frac{1}{\sqrt{m}}\bar{A}(x-x^*)}_{\overline{\bbY}} + \frac{1}{\sqrt{\alpha}}  \nms{\frac{1}{\sqrt{m}}\bar{A}(x-u^*)}_{\overline{\bbY}} + \nm{x - x^*}_{\bbX} + \frac{1}{\sqrt{\alpha}}  \nm{\bar{e}}_{\overline{\bbY}} + \frac{\gamma}{\sqrt{\alpha}} .
}
By the Cauchy--Schwarz inequality, we deduce that
\bes{
\bbE ( \nm{x - \check{x}}^2_{\bbX} | E ) \lesssim \frac{1}{\alpha} \bbE  \nms{\frac{1}{\sqrt{m}}\bar{A}(x-x^*)}^2_{\overline{\bbY}} + \bbE  \nms{\frac{1}{\sqrt{m}}\bar{A}(x-u^*)}^2_{\overline{\bbY}} + \nm{x - x^*}^2_{\bbX} + \frac{1}{\alpha} \nm{\bar{e}}^2_{\overline{\bbY}} + \frac{\gamma^2}{\alpha}.
}
We now use \ef{nondegeneracy-equiv} to deduce that
\be{
\label{first-exp-term}
\bbE ( \nm{x - \check{x}}^2_{\bbX} | E ) \lesssim  \frac{\beta}{\alpha}  \nm{x-x^*}^2_{\bbX} +  \frac{\beta}{\alpha}  \nm{x-u^*}^2_{\bbX} + \frac{1}{\alpha} \nm{\bar{e}}^2_{\overline{\bbY}} + \frac{\gamma^2}{\alpha}.
}
We next consider the second term of \ef{exp-split}. Using the properties of $\cC$, we see that $\nm{x - \check{x}}_{\bbX} \leq 2 \theta$.
Substituting this and \ef{first-exp-term} into \ef{exp-split} and recalling that $\bbP(E^c) \leq \epsilon$ now gives the result.

Finally, it remains to show that the result holds when \ef{main-res-meas-cond} is replaced by \ef{main-res-meas-cond-alt}. To do this, we show that the right-hand side of \ef{main-res-meas-cond-alt} is an upper bound for that of \ef{main-res-meas-cond} for a suitable value of $q$.
We first bound the empirical norm $\nms{\cdot}_{q,\bar{A}}$ in terms of $\tnm{\cdot}_{\bbX}$. Observe that
\bes{
\nm{A_i(u_1-u_2)}^2_{\bbY_i} \leq \Phi(S(\Delta \widetilde{S}(\bbU^*)) ; \bar{\cA} ) \nm{u_1-u_2}^2_{\bbX},\quad \forall u_1,u_2 \in \widetilde{S}(\bbU^*)
}
with probability one for $\bar{A} \sim \bar{\cA}$. Therefore
\bes{
\nm{u_1-u_2}_{q,\bar{A}} \leq \sqrt{\frac{\Phi(S(\Delta \widetilde{S}(\bbU^*)) ; \bar{\cA} )}{\alpha}} \tnm{u_1-u_2}_{\bbX},\quad \forall u_1,u_2 \in \widetilde{S}(\bbU^*)
}
with probability one. Hence, using properties (D) and (E) of \S \ref{ss:cov-num-prop}, we see that the integral in \ef{main-res-meas-cond} satisfies
\be{
\label{replace-emp-norm-by-X-norm}
\esssup_{\bar{A} \sim \bar{\cA}} \int^{1/2}_{0} \sqrt{\log(2 \cN(\widetilde{S}(\bbU^*) , \nms{\cdot}_{q,\bar{A}} , \upsilon t  ) ) } \D t \leq \int^{1/2}_{0} \sqrt{\log(2 \cN(\widetilde{S}(\bbU^*) , \tnm{\cdot}_{\bbX} , \tau t  ) ) } \D t
}
for any $q \geq 1$.
We now set $q = 1 + \log(1+v^2)$.  This gives
\be{
\label{upsilon-q-bd}
\upsilon^{1+\frac{1}{q}} \leq \upsilon (1+\upsilon^2)^{\frac{1}{2q}} \leq \upsilon (1+\upsilon^2)^{\frac{1}{2(\log(1+\upsilon^2))}} = \upsilon \sqrt{\E}.
}
The result now follows after recalling that $\upsilon^2 = \Phi(S(\bbU^*) ; \bar{\cA})$.
}

\rem{[On the error bounds in expectation]
\label{rem:on-error-bounds}
The above proof explains why the truncation term is needed: namely, it bounds the error in the event where empirical nondegeneracy \ef{empirical-nondegeneracy} fails. Modifying the estimator appears unavoidable if one is to obtain an error bound in expectation. A downside of the estimator $\check{x}$ is that it requires an a priori bound on $\nm{x}_{\bbX}$. In some limited scenarios, one can avoid this by using a different estimator \cite{adcock2025optimal2}. Indeed, let $E$ be the event in the above proof. Then define the conditional estimator $\check{x}$ by $\check{x} = \hat{x}$ if $E$ occurs and zero otherwise. One readily deduces that this estimator obeys the same error bound \ef{main-res-err-bd} with $\theta$ replaced by $\nm{x}_{\bbX}$. Unfortunately, computing this estimator involves computing the empirical nondegeneracy constants. This is possible in some limited scenarios, such as when $\bbU$ is a linear subspace, as the constants then correspond to the maximum and minimum singular values of a certain matrix. However, this is generally impossible in the nonlinear case. For instance, in the classical compressed problem, we previously noted that \ef{empirical-nondegeneracy} is equivalent to the RIP. It is well known that computing RIP constants in NP-hard \cite{tillmann2014computational}.
}

\rem{
The observant reader may have noticed a small technical issue with Theorem \ref{t:main-res} and its proof: the estimator $\hat{x}$ (and therefore $\check{x}$) is nonunique, and therefore $\nm{x - \check{x}}_{\bbX}$ is not a well-defined random variable. To resolve this issue, one can replace $\nm{x - \check{x}}_{\bbX} = \nm{x - \cC(\hat{x})}_{\bbX}$ by the maximum error between $x$ and \textit{any} $\gamma$-minimizer $\hat{x}$. The error bound remains the same for this (well-defined) random variable.
}

\section{Proof of Corollaries \ref{cor:main-res-deltaU-cone}, \ref{cor:main-res-deltaU-subspaces-I} and \ref{cor:main-res-deltaU-subspaces-II}}

We now prove the three main corollaries.

\prf{
[Proof of Corollary \ref{cor:main-res-deltaU-cone}]
We use Theorem \ref{t:main-res}. Let $x^* = u^* \in \bbU$ be any point such that $\nm{x-u^*}_{\bbX} \leq 2 \inf_{u \in \bbU} \nm{x-u}_{\bbX}$ (the constant $2$ is arbitrary). In this case, the error bound \ef{main-res-err-bd} reduces to \ef{main-res-deltaU-cone-err-bd}. Therefore, it suffices to show that \ef{main-res-deltaU-cone-meas-cond-0}--\ef{main-res-deltaU-cone-meas-cond-2} imply either \ef{main-res-meas-cond} or \ef{main-res-meas-cond-alt}.

\pbk
\textit{Step 1:  \ef{main-res-deltaU-cone-meas-cond-0} implies \ef{main-res-meas-cond}.}
We do this by upper bounding the right-hand side of \ef{main-res-deltaU-cone-meas-cond-0} by that of \ef{main-res-meas-cond}. First, observe that $x^* \in \bbU$ implies that $\bbU^* = \bbU - \{ x^* \} \subseteq \Delta \bbU$.
It follows that $\widetilde{S}(\bbU^*) \subseteq \widetilde{S}(\Delta\bbU)$.
Hence the right-hand side of \ef{main-res-meas-cond} can be bounded above by
\be{
\label{udem-construction}
\upsilon^{2+\frac2q} \cdot \left [ 
\left( \esssup_{\bar{A} \sim \bar{\cA}} \int^{1/2}_{0} \sqrt{\log(2 \cN(\widetilde{S}(\Delta\bbU) , \nms{\cdot}_{q,\bar{A}} , \upsilon t  /2 ) ) } \D t  \right)^2 
+ \log(2/\epsilon) \right ].
}
Now, since $\bbU^* \subseteq \Delta \bbU$, we see that $\upsilon^2 = \Phi(S(\bbU^*) ; \bar{\cA}) / \alpha \leq \Phi(S(\Delta \bbU) ; \bar{\cA}) / \alpha $.
Therefore, recalling that the variation can be replaced by any upper bound (Remark \ref{rem:var-upper-bounds}), we deduce that \ef{udem-construction} can be upper bounded by the right-hand side of \ef{main-res-deltaU-cone-meas-cond-0}. This gives the result.

\pbk
\textit{Step 2:  \ef{main-res-deltaU-cone-meas-cond-1} implies \ef{main-res-meas-cond-alt}.} Suppose that $\Delta \bbU$ is a cone. Then $\widetilde{S}(\bbU^*) \subseteq \widetilde{S}(\Delta \bbU) = \{ v / \tnm{v}_{\bbX} : v \in \Delta \bbU \} \subseteq \Delta \bbU$.
This implies that $S(\Delta \widetilde{S}(\bbU^*)) \subseteq S(\Delta^2 \bbU)$,
from which it follows that $\Phi (S(\widetilde{S}(\bbU^*) - \widetilde{S}(\bbU^*) ) ; \bar{\cA} ) \leq \Phi(S(\Delta^2 \bbU ) ; \bar{\cA})$
and therefore
\bes{
 \sqrt{\frac{\Phi(S(\bbU^*) ; \bar{\cA} )}{\Phi(S(\widetilde{S}(\bbU^*) - \widetilde{S}(\bbU^*)) ; \bar{\cA} )} } \geq  \sqrt{\frac{\Phi(S(\bbU^*) ; \bar{\cA} )}{\Phi(S(\Delta^2 \bbU)) ; \bar{\cA} )} } .
}
Since the covering number $\cN(\cdot , \cdot , t)$ decreases as $t$ increases, we see that the right-hand side of \ef{main-res-meas-cond-alt} can be upper bounded  (up to a constant) by
\be{
\label{udem-construction-2}
\alpha^{-1} \cdot \Phi(S(\bbU^*) ; \bar{\cA}) \cdot \left [ \left( \int^{1/2}_{0} \sqrt{\log(2 \cN(\widetilde{S}( \Delta \bbU) , \tnm{\cdot}_{\bbX} , \tau t  ) ) } \D t  \right)^2 + \log(2/\epsilon) \right ],
}
where $\tau = \sqrt{\Phi(S(\bbU^*) ; \bar{\cA} ) / \Phi(S(\Delta^2 \bbU)) ; \bar{\cA} ) }$.
Next, using the fact that $\bbU^* \subseteq \Delta \bbU$, we see that $\Phi(S(\bbU^*) ; \bar{\cA}) \leq \Phi(S(\Delta \bbU) ; \bar{\cA})$.
Therefore, using Remark \ref{rem:var-upper-bounds} once more, we deduce that \ef{udem-construction-2} can be upper bounded by the right-hand side of \ef{main-res-deltaU-cone-meas-cond-1}.

\pbk
\textit{Step 3:  \ef{main-res-deltaU-cone-meas-cond-2} implies \ef{main-res-meas-cond-alt}.} It suffices to show that \ef{main-res-deltaU-cone-meas-cond-2} implies \ef{main-res-deltaU-cone-meas-cond-1}. Since $\Delta \bbU$ is a cone, we have $0 \in \Delta \bbU$. Therefore $\Delta \bbU \subseteq \Delta^2 \bbU$. It follows that $\Phi(\Delta \bbU ; \bar{\cA}) \leq \Phi(\Delta^2 \bbU ; \bar{\cA})$.
Hence, using Remark \ref{rem:var-upper-bounds} again, we may replace $\Phi(\Delta \bbU ; \bar{\cA})$ by the upper bound $\Phi(\Delta^2 \bbU ; \bar{\cA})$ in \ef{main-res-deltaU-cone-meas-cond-1}. In particular, the term $\tau$ can be replaced by $1$. This immediately leads to \ef{main-res-deltaU-cone-meas-cond-2}.
}

The proof of Corollary \ref{cor:main-res-deltaU-subspaces-I} modifies the proof given in \cite{adcock2024unified}. We first require the following.

\lem{
\label{lem:cov-num-int-uofs-bd}
Let $\bbU$ be as in Corollary \ref{cor:main-res-deltaU-subspaces-I}. Then
\bes{
\left ( \int^{1/2}_{0} \sqrt{\log(2 \cN (\widetilde{S}(\Delta \bbU) , \tnm{\cdot}_{\bbX} , \upsilon t  ) ) } \D t \right )^2 \lesssim \log(2 d) + n \log(2(1+\upsilon^{-1})),\quad \forall \upsilon > 0.
}
}
\prf{
We follow the arguments given in the proof of Lemma E.6 of \cite{adcock2024unified}. Assumption (ii) of Corollary \ref{cor:main-res-deltaU-subspaces-I} implies that $\widetilde{S}(\Delta\bbU) \subseteq \widetilde{B}(\bbV_1) \cup \cdots \cup \widetilde{B}(\bbV_d)$, where $\widetilde{B}(\bbV_i) = \{ v_i \in \bbV_i : \tnm{v_i}_{\bbX} \leq 1 \}$ is the unit ball of $(\bbV_i,\tnm{\cdot}_{\bbX})$. Using properties (A), (B) and (F) of \S \ref{ss:cov-num-prop}, we get
\eas{
\cN(\widetilde{S}(\Delta\bbU),\tnm{\cdot}_{\bbX}, \upsilon t) & \leq \cN(\widetilde{B}(\bbV_1) \cup \cdots \cup \widetilde{B}(\bbV_d) , \tnm{\cdot}_{\bbX}, \upsilon t / 2)
\\
 & \leq \sum^{d}_{i=1} \cN(\widetilde{B}(\bbV_i), \tnm{\cdot}_{\bbX}  , \upsilon t / 2)
 \leq d \left (1+8 (\upsilon t)^{-1} \right )^n.
}
Note that in the final step when we use property (F) we use the fact that $(\bbV_i , \tnm{\cdot}_{\bbX})$ is isometric to $(\bbR^{2n} , \nms{\cdot}_{\ell^2})$ in general (since we consider Hilbert spaces over complex numbers).
Therefore
\eas{
\left ( \int^{1/2}_{0} \sqrt{\log(2 \cN (\widetilde{S}(\Delta \bbU) , \tnm{\cdot}_{\bbX} , \upsilon t  ) ) } \D t \right )^2  & \lesssim \log(2 d) + n \left ( \int^{1/2}_{0} \sqrt{\log(1+8 (\upsilon t)^{-1}) } \D t \right )^2
\\
& \lesssim \log(2 d) + n \log(\E(1+8 \upsilon^{-1})),
}
where in the second step we used \ef{log-integral-bd}.
}

\prf{
[Proof of Corollary \ref{cor:main-res-deltaU-subspaces-I}]
We divide the proof into three cases, depending on the measurement condition (a), (b) or (c). For condition (a), we also split into two cases depending on which term attains the minimum.

\pbk
\textit{Condition (a) when $\Phi(S(\Delta^2 \bbU ) ; \bar{\cA})$ attains the minimum.} 
We use Corollary \ref{cor:main-res-deltaU-cone} and, specifically, we show that condition (a) implies  \ef{main-res-deltaU-cone-meas-cond-1}. Let $\tau = \sqrt{\frac{\Phi(S(\Delta \bbU) ; \bar{\cA} )}{\Phi(S(\Delta^2 \bbU)) ; \bar{\cA} ) } }$ be as in Corollary \ref{cor:main-res-deltaU-cone}. By the previous lemma, we have
\eas{
\left ( \int^{1/2}_{0} \sqrt{\log(2 \cN (\widetilde{S}(\Delta \bbU) , \tnm{\cdot}_{\bbX} , \upsilon t  ) ) } \D t \right )^2 \lesssim \log(2 d) + n \log(\E(1+\upsilon^{-1})).
} 
Observe that $\tau \leq 1$, since $\Delta \bbU$ is a cone and therefore $\Delta \bbU \subseteq \Delta^2 \bbU$. Hence
\bes{
\left ( \int^{1/2}_{0} \sqrt{\log(2 \cN (\widetilde{S}(\Delta \bbU) , \tnm{\cdot}_{\bbX} , \upsilon t  ) ) } \D t \right )^2 \lesssim \log(2 d) + n \log \left ( 2 \frac{\Phi(S(\Delta^2 \bbU) ; \bar{\cA} )}{\Phi(S(\Delta \bbU) ; \bar{\cA} )} \right ).
}
We deduce that the right-hand side of \ef{main-res-deltaU-cone-meas-cond-1} is bounded above (up to a constant) by
\bes{
\alpha^{-1} \cdot \Phi(S(\Delta \bbU) ; \bar{\cA}) \cdot \left [ n \cdot \log \left ( 2 \frac{\Phi(S(\Delta^2 \bbU) ; \bar{\cA} )}{\Phi(S(\Delta \bbU) ; \bar{\cA} ) } \right ) + \log(2 d / \varepsilon ) \right ].
}
This is precisely the right-hand side of (a). This gives the result.

\pbk
\textit{Condition (a) when $\Phi(S(\bbV) ; \bar\cA)$ attains the minimum.} We use Corollary \ref{cor:main-res-deltaU-cone} and, specifically, the condition \ef{main-res-deltaU-cone-meas-cond-0}. 
As in the previous lemma, we write $\widetilde{S}(\Delta\bbU) \subseteq \widetilde{B}(\bbV_1) \cup \cdots \cup \widetilde{B}(\bbV_d)$, where $\widetilde{B}(\bbV_i) = \{ v_i \in \bbV_i : \tnm{v_i}_{\bbX} \leq 1 \}$ is the unit ball of $(\bbV_i,\tnm{\cdot}_{\bbX})$. Let $\bar{A} \sim \bar{\cA}$ and consider the norm $\nms{\cdot}_{q,\bar{A}}$ defined in \ef{discrete-norm-dudley-def}.
Then we have
\bes{
\cN(\widetilde{S}(\Delta \bbU) , \nms{\cdot}_{q,\bar{A}} , \upsilon t ) \leq \sum^{d}_{i=1} \cN(\widetilde{B}(\bbV_i),  \nms{\cdot}_{q,\bar{A}} , \upsilon t )
}
Now, since each $\bbV_i$ is a subspace, we have
\bes{
\nm{A_j(v-v')}^2_{\bbY_i} \leq \Phi(S(\bbV_i) ; \cA_j) \nm{v-v'}^2_{\bbX} \leq \Phi(S(\bbV) ; \bar{\cA}) \nm{v-v'}^2_{\bbX},\quad \forall v,v' \in \bbV_i,
}
with probability one for $\bar{A} \sim \bar{\cA}$. It follows that
\bes{
\nm{v-v'}_{q,\bar{A}} \leq \sqrt{\frac{\Phi(S(\bbV) ; \bar{\cA}) }{\alpha}} \tnm{v-v'}_{\bbX},\quad \forall v , v' \in \bbV,
}
with probability one.
Using properties (A), (B) and (F) of \S \ref{ss:cov-num-prop} once more, we deduce that 
\eas{
\cN(\widetilde{S}(\Delta \bbU) , \nms{\cdot}_{q,\bar{A}} , \upsilon t ) & \leq \sum^{d}_{i=1} \cN \left (\widetilde{B}(\bbV_i),  \tnm{\cdot}_{\bbX} , \upsilon t  \sqrt{\alpha / \Phi(S(\bbV) ; \bar{\cA})} /2 \right )
\\
& \leq d \left (1 + 8 \sqrt{ \frac{\Phi(S(\bbV) ; \bar{\cA})}{\alpha} } (\upsilon t)^{-1} \right )^n
}
and therefore 
\be{
\label{N-V-bd}
\cN(\widetilde{S}(\Delta \bbU) , \nms{\cdot}_{q,\bar{A}} , \upsilon t )  \leq d \left (1 + 8 \sqrt{ \frac{\Phi(S(\bbV) ; \bar{\cA})}{\Phi(S(\Delta \bbU) ; \bar{\cA} ) } } t^{-1} \right )^n
}
for any $t > 0$, where in the final step we used the definition of $\upsilon$ from Corollary \ref{cor:main-res-deltaU-cone}. Now consider the right-hand side of \ef{main-res-deltaU-cone-meas-cond-0}. Using the above arguments, we see that this is bounded (up to a constant) by
\bes{
\alpha^{-1} \cdot \Phi(S(\Delta \bbU) ; \bar{\cA}) \cdot \left [ \log(2 d/\varepsilon) + n \left ( \int^{1/2}_{0} \sqrt{\log \left ( 1 + 8 \sqrt{\frac{ \Phi(S(\bbV) ; \bar{\cA})}{ \Phi(S(\Delta \bbU) ; \bar{\cA})} } t^{-1} \right )} \D t \right )^2 \right ].
}
We now apply \ef{log-integral-bd} to see that this is bounded (up to a constant) by
\bes{
\alpha^{-1} \cdot \Phi(S(\Delta \bbU) ; \bar{\cA}) \cdot \left [ \log(2 d/\varepsilon) + n \cdot \log \left (2 \frac{ \Phi(S(\bbV) ; \bar{\cA})}{ \Phi(S(\Delta \bbU) ; \bar{\cA})} \right ) \right ] .
}
Here we also used the fact that $\Phi(S(\bbV) ; \bar{\cA}) \geq \Phi(S(\Delta \bbU) ; \bar{\cA})$, since $\Delta \bbU \subseteq \bbV$. This gives the result.

\pbk
\textit{Condition (b).} Condition (b) follows immediately from condition (a), after recalling that $\Phi(S(\Delta \bbU ) ; \bar{\cA}) \leq \Phi( S(\Delta^2 \bbU) ; \bar{\cA})$ since $\Delta \bbU \subseteq \Delta^2 \bbU$ and then using Remark \ref{rem:var-upper-bounds}.

\pbk
\textit{Condition (c).} The proof under this condition was shown in \cite{adcock2024unified} (see the proof of Theorem E.3 therein). For succinctness, we omit it.
}

\lem{
\label{lem:maurey}
Suppose that $\bbU$ satisfies the assumptions of Corollary \ref{cor:main-res-deltaU-subspaces-II} and consider the norm \ef{discrete-norm-dudley-def}. Then
\bes{
\sqrt{\log(\cN(\widetilde{S}(\Delta \bbU) , \nms{\cdot}_{q,\bar{A}} , t )) } \lesssim \sqrt{\frac{q  \Phi(\bbW ; \bar{\cA}) \log(M) }{\alpha} } \cdot t^{-1}
}
with probability one for $\bar{A} \sim \bar{\cA}$.
}
\prf{
The proof is a minor modification of that of \cite[Lem.\ E.9]{adcock2024unified}. We therefore provide only the salient details. Since $\widetilde{S}(\bbU) \subseteq \mathrm{conv}(\bbW / \sqrt{\alpha})$, Maurey's lemma (see, e.g., \cite[Lem.\ 13.30]{adcock2021compressive}) gives
\bes{
\cN(\widetilde{S}(\Delta \bbU) , \nms{\cdot}_{q,\bar{A}} , t )) \leq \cN(\mathrm{conv}(\bbW / \sqrt{\alpha}) , \nms{\cdot}_{q,\bar{\cA}} , t/2) \lesssim (C/t)^2 \log(M),
}
where $C$ is such that
\bes{
\bbE_{\bar{\epsilon}} \nms{\sum^{L}_{l=1} \epsilon_l \frac{w_l}{\sqrt{\alpha}} }_{q,\bar{A}} \leq C \sqrt{L},\quad \forall L \in \bbN,\ w_1,\ldots,w_L \in \bbW.
}
Let $L \in \bbN$ and $w_1,\ldots,w_L \in \bbW$. By H\"older's inequality and linearity,
\bes{
\bbE_{\bar{\epsilon}} \nms{\sum^{L}_{l=1} \epsilon_l \frac{w_l}{\sqrt{\alpha}} }_{q,\bar{A}} \leq \frac{1}{\sqrt{\alpha}} \max_{i =1,\ldots,m} \left ( \bbE_{\bar{\epsilon}} \nms{\sum^{L}_{l=1} \epsilon_l A_i(w_l) }^{2q}_{\bbY_i} \right )^{\frac{1}{2q}}
}
and by \cite[Lem.\ E.8]{adcock2024unified} we have
\bes{
 \left ( \bbE_{\bar{\epsilon}} \nms{\sum^{L}_{l=1} \epsilon_l A_i(w_l) }^{2q}_{\bbY_i} \right )^{\frac{1}{2q}} \lesssim \sqrt{q L} \max_{l=1,\ldots,L} \nm{A_i(w_l)}_{\bbY_i} \leq \sqrt{q L \Phi(\bbW ; \bar{\cA} )},
}
with probability one for $\bar{A} \sim \bar{\cA}$. The result now follows.
}

\prf{[Proof of Corollary \ref{cor:main-res-deltaU-subspaces-II}]
We treat each of the three conditions separately. 

\pbk
\textit{Condition (a).}
We use Corollary \ref{cor:main-res-deltaU-cone} and, in particular, \ef{main-res-deltaU-cone-meas-cond-0}. Recalling Remark \ref{rem:var-upper-bounds}, we now replace $\upsilon$ defined therein by the upper bound
\be{
\label{new-upsilon}
\upsilon = \sqrt{\Phi(S(\Delta \bbU) \cup \bbW ; \bar{\cA} ) / \alpha} \geq \sqrt{\Phi(S(\Delta \bbU) ; \bar{\cA} ) / \alpha} 
}
(we continue to label it as $\upsilon$ for convenience).
Now consider the integral in \ef{main-res-deltaU-cone-meas-cond-0} and let $0 < \theta < 1/2$ and $q \geq 1$ (we will choose their values later). Then we split the integral as
\be{
\label{I-barA-def}
I_{\bar{A}} : =  \int^{1/2}_{0} \sqrt{\log(2 \cN(\widetilde{S}(\Delta\bbU) , \nms{\cdot}_{q,\bar{A}} , \upsilon t  ) ) } \D t  = I_{1,\bar{A}} + I_{2,\bar{A}},
}
where 
\be{
\label{I-barA-split}
\begin{split}
I_{1,\bar{A}} &=  \int^{\theta}_{0} \sqrt{\log(2 \cN(\widetilde{S}(\Delta\bbU) , \nms{\cdot}_{q,\bar{A}} , \upsilon t  ) ) } \D t ,\\ I_{2,\bar{A}} & =  \int^{1/2}_{\theta} \sqrt{\log(2 \cN(\widetilde{S}(\Delta\bbU) , \nms{\cdot}_{q,\bar{A}} , \upsilon t  ) ) } \D t . 
\end{split} 
}
We first bound $I_{1,\bar{A}}$. We do this in two cases, depending on which term attains the minimum in condition (a). Suppose first that $ \Phi(S(\Delta^2 \bbU) ; \bar{\cA} ) $ attains the minimum. Then, by essentially the same argument that leads to \ef{replace-emp-norm-by-X-norm}, we deduce that 
\bes{
\esssup_{\bar{A} \sim \bar{\cA}} I_{1,\bar{A}} \leq 2 \theta \int^{1/2}_{0} \sqrt{\log(2 \cN(\widetilde{S}(\Delta \bbU) , \tnm{\cdot}_{\bbX} , 2 \theta \tau t  ) ) } \D t,
}
where $\tau = \sqrt{\frac{\Phi(S(\Delta \bbU) \cup \bbW  ; \bar{\cA} )}{\Phi(S(\Delta^2 \bbU)) ; \bar{\cA} ) } }$. We now apply Lemma \ref{lem:cov-num-int-uofs-bd} to see that
\be{
\label{I1A-bd}
\esssup_{\bar{A} \sim \bar{\cA}} I_{1,\bar{A}} \lesssim \theta \sqrt{\log(2 d) + n \log(\theta^{-1} (1+\tau^{-2})) },
}
where we also used the fact that $0 < \theta < 1/2$. Conversely, suppose that $\Phi(S(\bbV) ; \bar{\cA} )$ attains the minimum. We use \ef{N-V-bd}, \ef{log-integral-bd} and some algebraic manipulations to write
\eas{
\esssup_{\bar{A} \sim \bar{\cA}} I_{1,\bar{A}} & \leq \int^{\theta}_{0} \sqrt{\log(d)} + \sqrt{n} \sqrt{\log  \left (1 + 4 \tau^{-1} t^{-1} \right ) } \D t
\\
& = \theta \sqrt{\log(d)} + 2 \theta \sqrt{n} \int^{1/2}_{0} \sqrt{\log  \left (1 + 2 \theta^{-1} \tau^{-1}  t^{-1} \right ) } \D t
\\
& \lesssim \theta \left ( \sqrt{\log(2d)}  + \sqrt{n} \sqrt{\log \left ( \theta^{-1} (1+\tau^{-2} \right )} \right )
}
where $\tau = \sqrt{ \frac{\Phi(S(\Delta \bbU) \cup \bbW; \bar{\cA} ) }{\Phi(S(\bbV) ; \bar{\cA})} }$. Hence, in general, we have
\bes{
\esssup_{\bar{A} \sim \bar{\cA}} I_{1,\bar{A}} \lesssim  \theta \left ( \sqrt{\log(2d)}  + \sqrt{n} \sqrt{\log \left ( \theta^{-1} (1+\tau^{-2} \right )} \right )
}
where $\tau = \sqrt{ \frac{\Phi(S(\Delta \bbU) \cup \bbW; \bar{\cA} ) }{\min \{ \Phi(S(\Delta^2 \bbU)) ; \bar{\cA} )  ,  \Phi(S(\bbV) ; \bar{\cA}) \} } }$.

We now consider $I_{2,\bar{A}}$. By Lemma \ref{lem:maurey}, we get
\bes{
\esssup_{\bar{A} \sim \bar{\cA}} I_{2,\bar{A}} \lesssim \sqrt{\frac{q \Phi(\bbW ; \bar{\cA}) \log(M) }{\upsilon^2 \alpha} } \log(1/\theta).
}
Combining this with the bound for $I_{1,\bar{A}}$ and using \ef{new-upsilon}, we deduce that
\eas{
\esssup_{\bar{A} \sim \bar{\cA}} I_{\bar{A}} & \lesssim \theta \sqrt{\log(2 d) + n \log(\theta^{-1} (1+ \tau^{-2}) ) } + \sqrt{q  \log(M) } \log(1/\theta)
\\
& \lesssim \theta \sqrt{ \log(2d) + n } \sqrt{\log(\theta^{-1} (1+ \tau^{-2}))}+ \sqrt{q  \log(M) } \log(1/\theta).
}
Now set $\theta = (4(\log(2d) + n))^{-1/2}$, to obtain
\bes{
\esssup_{\bar{A} \sim \bar{\cA}} I_{\bar{A}} \lesssim \sqrt{L_1} + \sqrt{q} \sqrt{L_2},
}
where 
\eas{
L_1 &= \log \left [ 2 (\log(2d) + n) \right ] + \log \left (1 + \frac{\min \{ \Phi(S(\Delta^2 \bbU)) ; \bar{\cA} )  ,  \Phi(S(\bbV) ; \bar{\cA}) \}}{\Phi(S(\Delta \bbU) \cup \bbW ; \bar{\cA} )}  \right )
\\
L_2  &= \log(M) \cdot \log^2 \left [ 2 (\log(2d)+n) \right ].
}
Finally, we now set $q = 1 + \log(1+\upsilon^2)$. Using \ef{upsilon-q-bd}, \ef{new-upsilon} and the previous inequality, we deduce that the right-hand side of \ef{main-res-deltaU-cone-meas-cond-0} can be bounded above by
\bes{
\alpha^{-1} \cdot \Phi(S(\Delta \bbU) \cup \bbW ; \bar{\cA} ) \cdot \left [ L_1 + \log(2+\Phi(S(\Delta \bbU) \cup \bbW ; \bar{\cA} ) / \alpha ) L_2 \right ].
}
After some straightforward algebraic manipulations, we obtain the result.

\pbk
\textit{Condition (b).} Notice that $\Phi(S(\Delta^2 \bbU)  ; \bar{\cA}) \leq  \Phi(S(\Delta^2 \bbU) \cup \bbW ; \bar{\cA})$ and $\Phi(S(\Delta \bbU) \cup \bbW ; \bar{\cA}) \leq \Phi(S(\Delta^2 \bbU) \cup \bbW ; \bar{\cA})$.
Using this and the inequality $\log(x) \leq x$ we deduce that
\bes{
\Phi(S(\Delta \bbU) \cup \bbW ; \bar{\cA} )  \cdot \log\left (1+  \frac{\Phi(S(\Delta^2 \bbU)) ; \bar{\cA} ) }{\Phi(S(\Delta \bbU) \cup \bbW ; \bar{\cA} )} \right ) \leq 2 \Phi(S(\Delta^2 \bbU) \cup \bbW ; \bar{\cA}).
}
From this we deduce that condition (b) implies condition (a). This gives the result.

\pbk
\textit{Condition (c).} Since $\Delta \bbU \subseteq \bbV$, we have $\Phi(S(\Delta \bbU) \cup \bbW) ; \bar{\cA}) \leq \Phi(S(\bbV) \cup \bbW) ; \bar{\cA})$
and therefore, by the same reasoning,
\bes{
\Phi(S(\Delta \bbU) \cup \bbW ; \bar{\cA} )  \cdot \log\left (1+  \frac{\Phi(S(\bbV) ; \bar{\cA} ) }{\Phi(S(\Delta \bbU) \cup \bbW ; \bar{\cA} )} \right ) \leq 2 \Phi(S(\bbV) \cup \bbW ; \bar{\cA}).
}
Hence condition (c) also implies condition (a).
}

\section{Proof of Theorem \ref{thm:lipschitz-model-classes} }

The proof of this result uses Theorem \ref{t:main-res} and, specifically, \ef{main-res-meas-cond-alt}. To prove it, we require a series of technical lemmas. The first two lemmas estimate the covering number in \ef{main-res-meas-cond-alt}. The third lemma bounds the error term \ef{main-res-err-bd} over the restricted range space $\bbU$.

\lem{
[Covering numbers of images of balls under Lipschitz maps]
\label{l:cov-num-lip-ball}
Let $\abs{\cdot}$ and $\nms{\cdot}$ be norms on $\bbR^k$ and $\bbR^N$, respectively, $F : (\bbR^k , \abs{\cdot}) \rightarrow (\bbR^N,\nms{\cdot})$ be Lipschitz with constant $L \geq 0$ and, for $r > 0$, let $B_r = \{ z \in \bbR^k : \abs{z} \leq r \}$. Then
\bes{
\cN(F(B_r) , \nms{\cdot} , t) \leq \left ( 1 + \frac{2L r}{t} \right )^k,\quad \forall t > 0.
}
}
\prf{
Let $z_1,\ldots,z_n$ be a $(t/L)$-covering of $B_r$. Let $x = F(z) \in F(B_r)$ and $i \in [n]$ be such that $\abs{z - z_i} \leq t/L$. Then $\nm{x - F(z_i)} = \nm{F(z) - F(z_i)}  \leq t$ and hence $F(z_1),\ldots,F(z_n)$ is a $t$-covering of $F(B_r)$. We deduce that $\cN(F(B_r) , \nms{\cdot}, t) \leq \cN( B_r , \abs{\cdot} , t/L)$.
The result now follows from property (F) of \S \ref{ss:cov-num-prop}.
}

Recall that in \S \ref{s:applic-gen-CS} we consider $\bbX = \bbR^N$ with the Euclidean norm. In the remainder of this section, we write $\tnm{\cdot}$ for the norm \ef{equiv-X-norm} induced by the sampling operators.

\lem{
[Covering number of restricted range spaces]
\label{lem:lipschitz-cover-num}
Consider the setup of Theorem \ref{thm:lipschitz-model-classes} and let $\bbU^* = \bbU - \{x^* \}$, where $x^*$ is as in (a). Then
\bes{
\cN(\widetilde{S}(\bbU^*) , \tnm{\cdot} , t ) \leq \left ( 1 + 
\frac{4L \eta}{\sqrt{\alpha} \chi \iota_x t}  \right )^{2k}.
}
}
\prf{
By (a) and (b) of Theorem \ref{thm:lipschitz-model-classes}, we have $\nm{x^*}_{\ell^2} \leq (1+\chi) \nm{x}_{\ell^2} \leq \frac{1+\chi}{1-\chi} \iota_x$.
This implies that
\bes{
\bbU^* \subseteq \left \{ u - x^* : u = F(z),\ \nm{z}_{\ell^2} \leq 2\eta ,\ \nm{u}_{\ell^2} \geq \frac{1+\chi}{1-\chi} \nm{x^*}_{\ell^2} \right \}.
}
Now define $G : \bbR^k \times \bbR^k \rightarrow \bbR^N$ by $G(z,z') = F(z) - F(z')$. Then (ii) of Theorem \ref{thm:lipschitz-model-classes} implies that $G$ is positively homogeneous, and therefore
\eas{
\widetilde{S}(\bbU^*) & \subseteq \left \{ \frac{G(z,z^*)}{\tnm{G(z,z^*)}} : \nm{z}_{\ell^2} \leq 2 \eta ,\ \nm{F(z)}_{\ell^2} \geq \frac{1+\chi}{1-\chi} \nm{x^*}_{\ell^2} \right \}
\\
& = \left \{ G \left ( \frac{z}{\tnm{G(z,z^*)}} , \frac{z^*}{\tnm{G(z,z^*)}} \right ) : \nm{z}_{\ell^2} \leq 2\eta ,\ \nm{F(z)}_{\ell^2} \geq \frac{1+\chi}{1-\chi} \nm{x^*}_{\ell^2} \right \}.
}
Now observe that, by (a),
\eas{
\nms{ \frac{z}{\tnm{G(z,z^*)}} }_{\ell^2} + \nms{ \frac{z^*}{\tnm{G(z,z^*)}} }_{\ell^2} \leq \frac{\nm{z}_{\ell^2} + \nm{z^*}_{\ell^2}}{\sqrt{\alpha} \nm{G(z,z^*)}_{\ell^2}}
& \leq \frac{2\eta + \eta}{\sqrt{\alpha}(\nm{F(z)}_{\ell^2} - \nm{F(z^*)}_{\ell^2} )}
\\
& = \frac{3 \eta }{\sqrt{\alpha}\left ( \frac{1+\chi}{1-\chi} \nm{x^*}_{\ell^2} - \nm{x^*}_{\ell^2} \right ) }
\\
& \leq \frac{3 \eta }{\sqrt{\alpha} \frac{2 \chi}{1-\chi} \nm{x^*}_{\ell^2}},
}
and therefore, by (a) and (b) and the fact that $\chi \leq 1/3$,
\eas{
\nms{ \frac{z}{\tnm{G(z,z^*)}} }_{\ell^2} + \nms{ \frac{z^*}{\tnm{G(z,z^*)}} }_{\ell^2}  & \leq \frac{3 \eta }{\sqrt{\alpha} \frac{2\chi}{1-\chi} \frac{1-\chi}{1+\chi} \iota_x} \leq \frac{2 \eta}{\sqrt{\alpha} \chi \iota_x}. 
}
We deduce that
\bes{
\widetilde{S}(\bbU^*) \subseteq \left \{ G(z,z') : z,z' \in \bbR^k,\ \nm{z}_{\ell^2} + \nm{z'}_{\ell^2} \leq \frac{2\eta }{\sqrt{\alpha} \chi \iota_x} \right \}.
}
The set on the right-hand side is the image under $G$ of the ball centred at the origin of radius $\frac{2\eta }{\sqrt{\alpha} \chi \iota_x}$ with respect to the norm $\nm{(z,z')} = \nm{z}_2 + \nm{z'}_2$ on $\bbR^{2k}$. The map $G$ is Lipschitz since $F$ is, and has the same Lipschitz constant with respect to this norm. The result now follows from Lemma \ref{l:cov-num-lip-ball}.
}

As discussed, when using a Lipschitz map $F$ one normally strives to approximate an unknown target object $x$ from elements of $\mathrm{ran}(F)$. However, in order to obtain a desirable measurement condition, we need to restrict the model class as in \ef{U-Lipschitz-map-def}. The next lemma provides a best approximation error bound for this space.

\lem{
[Best approximation in the restricted range space]
\label{l:U-Lipschitz-closeness}
Consider the setup of Theorem \ref{thm:lipschitz-model-classes}. Then $\inf_{u \in \bbU} \nm{x - u}_{\ell^2} \lesssim \chi \nm{x}_{\ell^2}$.
}
\prf{
Let $\epsilon >0$ (its value will be chosen later) and define $x^*_{\epsilon} = (1+\epsilon) x^*$. Since $F$ satisfies (ii) of Theorem \ref{thm:lipschitz-model-classes}, we have $x^*_{\epsilon} = F(z^*_{\epsilon})$, where $z^*_{\epsilon} = (1+\epsilon) z^*$. By (a),
\bes{
\nm{z^*_{\epsilon}}_{\ell^2} = (1+\epsilon) \nm{z^*}_{\ell^2} \leq (1+\epsilon) \eta,
}
and, by (a) and (b),
\bes{
\nm{x^*_{\epsilon}}_{\ell^2} = (1+\epsilon) \nm{x^*}_{\ell^2} \geq (1+\epsilon)(1-\chi) \nm{x}_{\ell^2} \geq (1+\epsilon) \frac{1-\chi}{1+\chi} \iota_x.
}
Therefore, $x^*_{\epsilon} \in \bbU$, provided
\bes{
1+\epsilon \leq 2,\qquad (1+\epsilon) \frac{1-\chi}{1+\chi} \geq \left (\frac{1+\chi}{1-\chi} \right )^2.
}
Rearranging the second inequality, we see that these are equivalent to $g(\chi) \leq \epsilon \leq 1$, where $g(\chi) = \left (\frac{1+\chi}{1-\chi} \right )^3-1$.
Let $\epsilon = 9 \chi$. Since $0 \leq \chi \leq 1/9$ by assumption, we have $\epsilon \leq 1$. We also have $g(\chi) \leq 9 (g(1/9)-g(0)) \chi < 9 \chi = \epsilon$.
Hence this choice of $\epsilon$ is valid and therefore $x^*_{\epsilon} \in \bbU$. Using this and (a) once more, we deduce that
\bes{
\inf_{u \in \bbU} \nm{x - u}_{\ell^2} \leq \nm{x - x^*_{\epsilon} }_{\ell^2} \leq \nm{x - x^*}_{\ell^2} + \epsilon \nm{x^*}_{\ell^2} \lesssim \chi \nm{x}_{\ell^2} + \chi \nm{x^*}_{\ell^2} \lesssim \chi \nm{x}_{\ell^2}.
}
This completes the proof.
}

\prf{
[Proof of Theorem \ref{thm:lipschitz-model-classes}]
We use Theorem \ref{t:main-res} and, specifically, \ef{main-res-meas-cond-alt}.  Using Lemmas \ref{lem:lipschitz-cover-num} and \ref{lem:integral-bound}, we see that the right-hand side of \ef{main-res-meas-cond-alt} can be bounded above (up to a constant) by
\bes{
\alpha^{-1} \cdot \Phi(S(\bbU^*) ; \bar{\cA}) \cdot \left [ k \log \left ( 2 \left ( 1 + \frac{L \eta}{\sqrt{\alpha} \chi \iota_x  \tau} \right ) \right ) + \log(2/\epsilon) \right ],
}
where $\tau$ is as in Theorem \ref{t:main-res}. Since $\bbU^* \subseteq \Delta \bbU  \subseteq \Delta \bbU_0$ and $\bbU_0$ is a cone (this follows from (ii)), we have
\bes{
\Phi(S(\bbU^*) ; \bar{\cA}) \leq \Phi(S(\Delta \bbU_0) ; \bar{\cA}),\quad \Phi(S(\Delta(\widetilde{S}(\bbU^*) ) ) ; \bar{\cA} ) \leq \Phi(S(\Delta^2 \bbU_0) ; \bar{\cA}),
}
which, due to Remark \ref{rem:var-upper-bounds}, means that the right-hand side of \ef{main-res-meas-cond-alt} can be bounded by
\bes{
\alpha^{-1} \cdot \Phi(S(\Delta \bbU_0) ; \bar{\cA}) \cdot \left [ k \log \left ( 2 \left ( 1 + \frac{L \eta \Phi(S(\Delta^2 \bbU_0) ; \bar{\cA})}{\sqrt{\alpha} \chi \iota_x \Phi(S(\Delta \bbU_0) ; \bar{\cA}) }  \right ) \right ) + \log(2/\epsilon) \right ].
}
Therefore \ef{meas-cond-lipschitz-alt-1} implies \ef{main-res-meas-cond-alt}. Using Remark \ref{rem:var-upper-bounds} once more, we also see that \ef{meas-cond-lipschitz-alt-2} implies \ef{meas-cond-lipschitz-alt-1}, and therefore \ef{main-res-meas-cond-alt}, since $\Phi(S(\Delta \bbU_0) ; \bar{\cA}) \leq \Phi(S(\Delta^2 \bbU_0) ; \bar{\cA})$.

We now apply Theorem \ref{t:main-res} to deduce that 
\bes{
\bbE \nm{x - \check{x}}^2_{\ell^2} \lesssim \frac{\beta}{\alpha}  \left (  \nm{x - x^*}^2_{\ell^2} + \inf_{u \in \bbU} \nm{x - u}^2_{\ell^2} \right )+ \theta^2 \epsilon + \frac{\gamma^2}{\alpha} + \frac{\nm{e}^2_{\overline{\bbY}}}{\alpha},
}
where $\check{x} = \min \{ 1 , \theta/\nm{\hat{x}}_{\ell^2} \} \hat{x}$ and $\hat{x}$ is any $\gamma$-minimizer of \ef{least-squares-problem} with noisy measurements \ef{noisy-meas}. To complete the proof we use Lemma \ref{l:U-Lipschitz-closeness}.
}

\section{Proof of Corollaries \ref{cor:gen-mod-unitary-sub-bernoulli} and \ref{cor:gen-mod-sub-block}}\label{s:proofs-end}

\prf{
[Proof of Corollary \ref{cor:gen-mod-unitary-sub-bernoulli}]
Let $\bar\cA = \{ \cA_i \}^{N}_{i=1}$ be the distribution defined in Example \ref{ex:bernoulli-model} and consider case (i). Observe that $\cA_i$ is a constant distribution if and only if $\pi_i = 1/m$. Hence
\bes{
\Phi(S(\Delta \bbU_0) ; \bar\cA) = \max_{i : \pi_i < 1/m} \sup \left \{ \frac{N | u^*_i v |^2}{m \pi_i \nm{v}^2_{\ell^2}} : v \in \Delta \bbU_0,\ v \neq 0 \right \} =  \max_{i : \pi_i < 1/m} \frac{N c_{\pi} \sigma_i }{m \sigma_i} = \frac{N c_{\pi}}{m}.
}
Now observe that the function $\sum^{N}_{i=1} \min \{1/m,\sigma_i / c \}$ is nonincreasing in $c$ and is bounded above by $\nm{\sigma}^2_{\ell^2} / c$. Therefore $1 \leq \nm{\sigma}^2_{\ell^2} / c_{\pi}$, i.e., $c_{\pi} \leq \nm{\sigma}^2_{\ell^2}$. We deduce that
\bes{
\Phi(S(\Delta \bbU_0) ; \bar\cA) \leq \frac{N \nm{\sigma}^2_{\ell^2}}{m}.
}
Further, it is a short argument to see that
\bes{
\Phi(S(\Delta^2 \bbU_0) ; \bar\cA) \leq \frac{N \nm{\sigma}^2_{\ell^2}}{m} \max_{i=1,\ldots,N} \left \{ \frac{\tilde\sigma_i }{ \sigma_i} \right  \}.
}
Now recall that in this model $\bar{\cA}$ consists of $N$ distributions. Using this, the above bounds and the fact that $\alpha = 1$ for this model, we see that \ef{meas-cond-lipschitz-alt-1} is implied by the condition
\bes{
N \gtrsim \frac{N \nm{\sigma}^2_{\ell^2}}{m} \cdot \left [ k \cdot \log \left ( 2 \left ( 1 + \frac{L}{\lambda} \max_{i=1,\ldots,N} \left \{ \frac{\tilde\sigma_i }{ \sigma_i} \right  \} \right ) \right ) + \log(2/\epsilon) \right ].
}
Rearranging gives the result for case (i). Case (ii) is similar.
}

\prf{
[Proof of Corollary \ref{cor:gen-mod-sub-block}]
Let $\bar{\cA}$ be the distribution defined in \S \ref{ss:block-sampling}. Then
\eas{
\Phi(S(\bbV) ; \bar{\cA}) & = \max_{i=1,\ldots,M} \sup \left \{ \frac{\nm{B_i v}^2_{\ell^2} }{\nm{v}^2_{\ell^2}} : v \in \bbV,\ v \neq 0\right \}
\\
& = \max_{i=1,\ldots,M} \sup \left \{ \frac{\sum_{j \in P_i} | u^*_j v |^2 / r_j }{\pi_i \nm{v}^2_{\ell^2}} : v \in \bbV,\ v \neq 0\right \}  = \max_{i = 1,\ldots,M } \left \{ \frac{\sigma_i}{\pi_i} \right \},
}
where the $\sigma_i$ are as in \ef{sigma-i-def-block}. Hence, \ef{variation-sigmas} holds in this setting, with $M$ replaced by $N$ and the $\sigma_i$ given by \ef{sigma-i-def-block}. Since Corollaries \ref{cor:gen-mod-sub-unitary}--\ref{cor:gen-mod-unitary-sub-bernoulli} rely on plugging \ef{variation-sigmas} into the general result, Theorem \ref{thm:lipschitz-model-classes}, we immediately deduce that they all generalize to the block sampling model. 
}


\section*{Acknowledgements}

BA acknowledges support from the Natural Sciences and Engineering Research Council of Canada (NSERC) through grant RGPIN/2470-2021 and the support of FRQ (Fonds de recherche du Qu\'ebec) - Nature et Technologies through grant 359708.

\bibliographystyle{abbrv}
\small
\bibliography{RecovGuarGenModelsBib}

\end{document}